\documentclass{new_tlp}
\usepackage{mathptmx}

\usepackage{times,helvet,courier}
\usepackage{amsfonts}
\usepackage{csquotes}
\usepackage{caption}
\usepackage{listings}
\usepackage{color}
\usepackage{multirow,url}

\definecolor{dkgreen}{rgb}{0,0.6,0}
\definecolor{gray}{rgb}{0.5,0.5,0.5}
\definecolor{mauve}{rgb}{0.58,0,0.82}

\usepackage{amsmath}
\usepackage{algorithm}
\usepackage[noend]{algpseudocode}

\algnewcommand\algorithmicinput{\textbf{Input:}}
\algnewcommand\INPUT{\item[\algorithmicinput]}
\algnewcommand\algorithmicoutput{\textbf{Output:}}
\algnewcommand\OUTPUT{\item[\algorithmicoutput]}

\algblock{ParFor}{EndParFor}
\algnewcommand\algorithmicparfor{\textbf{do in parallel for}}
\algnewcommand\algorithmicendparfor{\textbf{end\ do}}
\algrenewtext{ParFor}[1]{\algorithmicparfor\ #1}
\algrenewtext{EndParFor}%{\algorithmicendparfor}

\algtext*{EndParFor}

\usepackage{xspace,epsfig}

\newcommand{\amp}[1]{\ensuremath{\text{\textsl{{\&}}}\!\,\mathit{#1}}}
\newcommand{\ext}[3]{\ensuremath{\amp{#1}[#2](#3)}}

\def\clingo{{\sc Clingo}\xspace}

\def\ccalc{{\sc CCalc}\xspace}
\def\cplus{${\cal C}+$\xspace}

\newcommand{\hex}{\textsc{hex}\xspace}
\def\openrave{{\sc OpenRave}}
\def\minisat{{\sc minisat}\xspace}

\def\lar{\leftarrow}
\def\ba{\begin{array}}
\def\ea{\end{array}}
\def\beq{\begin{equation}}
\def\eeq#1{\label{#1}\end{equation}}
\def\beqq{\begin{equation*}}
\def\eeqq{\end{equation*}}
\def\no{\ii{not}}
\def\ii#1{\hbox{\it #1\/}}

\newcommand{\ascp}{ASCP}
\def\openrave{{\sc OpenRave}}
\def\ompl{{\sc OMPL}}

\newcommand{\hcpasp}{HCP-ASP}

  \title[Hybrid Conditional Planning using ASP]
  {Hybrid Conditional Planning \\
  using Answer Set Programming}

  \author[Yalciner et al.]
  {Ibrahim Faruk Yalciner, Ahmed Nouman, Volkan Patoglu, and Esra Erdem\\
  Faculty of Engineering and Natural Sciences, Sabanci University, Istanbul, Turkey\\
  \email{{fyalciner,ahmednouman,vpatoglu,esraerdem}@sabanciuniv.edu}}

\jdate{March 2003}
\pubyear{2003}
\pagerange{\pageref{firstpage}--\pageref{lastpage}}
\doi{S1471068401001193}

\begin{document}

\label{firstpage}

\maketitle
\vspace{-1ex}
\begin{abstract}
We introduce a parallel offline algorithm for computing hybrid conditional plans, called \hcpasp, oriented towards robotics applications. \hcpasp\ relies on modeling actuation actions and sensing actions in an expressive nonmonotonic language of answer set programming (ASP), and computation of the branches of a conditional plan in parallel using an ASP solver.
In particular, thanks to external atoms, continuous feasibility checks (like collision checks) are embedded into formal representations of actuation actions and sensing actions in ASP; and thus each branch of a hybrid conditional plan describes a feasible execution of actions to reach their goals.
Utilizing nonmonotonic constructs and nondeterministic choices, partial knowledge about states and nondeterministic effects of sensing actions can be explicitly formalized in ASP; and thus each branch of a conditional plan can be computed by an ASP solver without necessitating a conformant planner and an ordering of sensing actions in advance.
We apply our method in a service robotics domain and report experimental evaluations. Furthermore, we present performance comparisons with other compilation based conditional planners on standardized benchmark domains. This paper is under consideration for acceptance in TPLP.
\end{abstract}

\begin{keywords}
Conditional planning, hybrid planning, answer set programming, cognitive robotics
\vspace{-2ex}
\end{keywords}
\vspace{-3ex}

%\tableofcontents

%%%%%%%%%%%%%%%%%%%%%%%%%%%%%%%%%%%%%%%%%%%%%%%%%%%%%%%%%%%%
\vspace{-1ex}
\section{Introduction} \label{sec:intro}

Conditional planning is concerned with reaching goals from an initial state, in the presence of incomplete knowledge and sensing actions~\cite{Warren1976,PeotS92,PryorC96}. Since all contingencies are considered in advance, a conditional plan is essentially a tree of actions where the root represents the initial state, leaves represent goal states, and each branch of the tree from the root to a leaf represents a possible execution of actuation actions and sensing actions to reach a goal state. The existence of a conditional plan is an intractable problem: for polynomially bounded plans with partial observability, it is PSPACE-complete~\cite{BaralKT99}.
Recall that classical planning considers complete knowledge and full observability, and it is NP-complete for polynomially bounded plans~\cite{ErolNS95}.

We are concerned about computing conditional plans that are executable by robots.
In real-life robotic applications, applicability of high-level planning
techniques (like navigation of a robot) necessitate low-level
feasibility checks (like collision checks). That a valid plan exists
does not mean that the plan is executable by a robot. Therefore,
some sort of hybrid planning is needed that combines planning with
feasibility checks. This is a challenging problem, considering that
high-level planning is done over discrete representations of the
world whereas feasibility checks are performed over continuous spaces of
robotic configurations.

Along these lines, several hybrid planning approaches
have been proposed in the literature, where classical planning is
integrated with low-level feasibility checks: by modifying the
search algorithms of the planners by embedding feasibility checks to
compute successor states, or by modifying the representations of the
domain descriptions by embedding feasibility checks into
preconditions (and other constraints) via semantic attachments.

In this paper, we consider both sorts of challenges, and introduce a
novel method for hybrid robotic planning under incomplete knowledge
that relies on the following contributions.

\begin{itemize}
\item
In conditional planning, in addition to familiar \emph{actuation}
actions, \emph{sensing} actions are also considered as part of the
planning problem, and different plans are computed according to all
possible outcomes of such sensing actions. As a result, a
conditional plan is a tree of  actions (Figure~\ref{fig:cplanTree}), where the branching occurs
at vertices that characterize sensing actions, while other vertices
characterize actuation actions.

\begin{figure}[htb]
  \vspace{-.5\baselineskip}
  \centering
  \resizebox{.45\columnwidth}{!}{\rotatebox{0}{\includegraphics{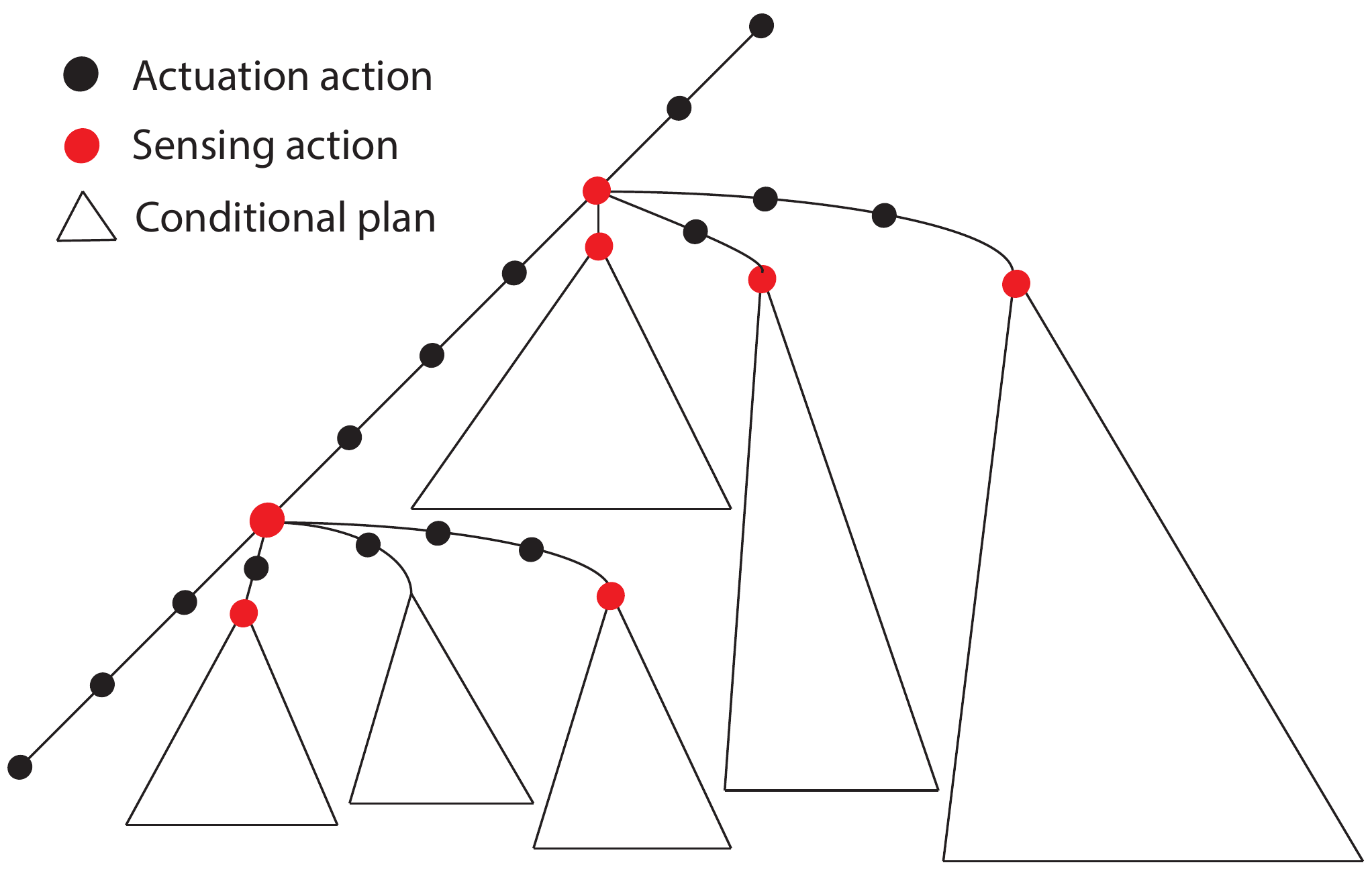}}}
  \vspace{-1\baselineskip}
  \caption{A sample conditional plan}
  \vspace{-\baselineskip}
  \label{fig:cplanTree}
\end{figure}

We introduce an offline compilation-based method to incrementally construct a conditional plan, that utilizes two sorts of parallelization.

\begin{itemize}
\item[(i)] Parts of branches (i.e., sequences of actuation and sensing actions that reach a goal state from a state where a sensing action occurs) are computed in parallel using classical planners and combined into subtrees as they are computed.

\item[(ii)] Subtrees of a conditional plan are computed in parallel according to (i) and then combined into larger subtrees. During this incremental construction, previously computed subtrees are re-used to improve computational efficiency.
\end{itemize}

\item
Since the existing classical planning approaches consider full observability and actuation actions only, they cannot be used in computation of branches in (i) which also involve partial observability and sensing actions.
To handle this challenge, we introduce a novel method to represent partial states (defined over functional fluents) and sensing actions using answer set programming (ASP) so that hybrid sequential plans of actuation actions and sensing actions can be computed using the existing ASP solvers.

\begin{itemize}
\item
Cardinality expressions are used to represent ``unknowns'' (e.g., when the value of a functional fluent is not known) in a partially observed state.

\item
Defaults are used  to further describe assumptions about partially observed worlds (e.g., unless the location of an object is known, the object is assumed to be not on the table since the table is fully observable).

\item
Defaults are used to describe exogenous occurrences (and non-occurrences) of sensing actions.
In this way, there is no need to decide on the order of sensing actions in advance (before conditional planning).

\item
Nondeterministic outcomes of sensing actions are represented using nondeterministic choices and cardinality constraints.
\end{itemize}

Since our method plans also for the sensing actions required to reach the goal in every branch of a conditional plan, unlike the existing compilation-based conditional planning methods, there is no need to decide on the order of sensing actions in advance (before conditional planning) or separate their computation from that of the actuation actions.
Also, since our method allows for modeling sensing actions as nondeterministic actions, unlike the existing compilation-based conditional planning methods, there is no need to use conformant planners to compute parts of branches that consist of actuation actions only and that occur between two sensing actions.

\item

To incrementally compute hybrid conditional plans for robotic applications as described above,
we adopt a representation-based approach to embed low-level feasibility
checks into action descriptions so that the
computed sequences of actions in (i), and thus conditional plans, are more likely to be physically implementable. In particular, we extend our representation-based approach to sensing actions as well.
Note that, since the existing representation-based approaches consider deterministic actuation actions only, they cannot be adopted for the computation of branches in (i) which also involve sensing actions.

\end{itemize}

We evaluate our method experimentally over a real-world service robotics domain where a bimanual mobile manipulator sets up a kitchen table in the presence of incomplete knowledge about kitchenware. Furthermore, we present performance comparisons with other compilation based conditional planners on standardized benchmark domains.

%%%%%%%%%%%%%%%%%%%%%%%%%%%%%%%%%%%%%%%%%%%%%%%%%%%%%%%%%%%%

\section{Related Work}\label{sec:related}

There is a variety of conditional planners, despite the hardness of computing conditional plans: even for polynomially bounded plans with limited number of sensing actions, its complexity is PSPACE-complete~\cite{BaralKT99}.  Some of these planners are online, such as CLG~\cite{AlborePG09}, K-Planner~\cite{BonetG11}, SDR~\cite{BrafmanS12}, HCP~\cite{MaliahBKS14} and CPOR~\cite{KomarnitskyS16}. Since sensing is done online, these solvers do not end up handling many contingencies and thus sometimes do not reach a goal state. Some of the conditional planners are offline, such as Contingent-FF~\cite{HoffmannB05}, POND~\cite{BryceKS06}, PKS~\cite{PetrickB02}, \ascp~\cite{TuSB07}, DNFct~\cite{ToSP11}, PO-PRP~\cite{MuiseBM14}, HCplan~\cite{NoumanYEP16}. These planners construct whole conditional plans with decision points for sensing outcomes, where each execution via a branch of the plan leads to a goal state. Our planner is an offline planner.

Offline conditional planners can be further classified into two groups: search-based approaches and compilation-based approaches. The former group views conditional planning as a nondeterministic search problem in belief space (e.g., Contingent-FF, PKS, POND), whereas most of the planners in the latter group compiles a conditional planning problem into many conformant planning problems by deciding for the order of sensing actions and then computing the branches of the conditional plan by means of conformant planning (e.g., CLG, DNFct, PO-PRP). Recall that conformant planning considers incomplete initial state and no observability, and its aim is to find an action sequence that reaches the goal for every possible initial state~\cite{Goldman1996}. Our planner can be characterized as a compilation-based planner since we compile a conditional planning problem into many sequential planning problems.

Essentially, we propose a generic hybrid conditional planning algorithm, called \hcpasp, with an offline compilation-based approach. Except for HCplan, it is different from  other offline compilation-based approaches, since it is hybrid (so low-level feasibility checks are integrated into conditional planning), it models multi-valued sensing actions as nondeterministic actions (so ordering of sensing actions and solving conformant planning problems are not needed as in the related work), and it allows for concurrency of actuation actions. See Table~\ref{tab:relatedWork} for a comparison.

\begin{table}[t]
%   \vspace*{-1\baselineskip}
\caption{A comparison of offline compilation-based conditional planners}
\label{tab:relatedWork}
{%\small
    %\begin{center}
        %\resizebox{\textwidth}{!}
        {
            \begin{tabular}{|c||r|r|r|r|}
                \cline{1-5}
                Name  &  Hybrid & Parallel  & Mode & Language (variant) \\
                \cline{1-5}
%               HCP & search & no & no & online & PDDL \\
%               Contingent-FF & search & no & no & offline & STRIPS \\
%               POND & search & no & no & offline & PPDDL \\
%               PKS & search & yes & no & offline & STRIPS \\
                CLG &  no & no & offline/online & PDDL\\
                \cline{1-5}
%               K-Planner &  no & no & online & STRIPS\\
%                \cline{1-5}
%               SDR &  no & no & online & PDDL \\
%               \cline{1-5}
                PO-PRP &  no & no & offline & PDDL/STRIPS \\
                \cline{1-5}
                DNFct &  no & no & offline & PDDL \\
                \cline{1-5}
                HCplan &  yes & yes & offline & $\cal C+$\\
                \cline{1-5}
                \textbf{\hcpasp} &  \textbf{yes} & \textbf{yes} & \textbf{offline} & \textbf {ASP}\\
                \cline{1-5}
            \end{tabular}}
        %\end{center}
        \vspace{-\baselineskip}
        }
    \end{table}

\hcpasp\ has similarities with HCplan: they both compute hybrid conditional plans, utilize parallel computation of branches, use nonmonotonic \ii{default} constructs to represent nonoccurrences of sensing actions, and outcomes of sensing actions are determined nondeterministically. They do not have to decide for the order of sensing actions in advance. On the other hand, the input language for HCplan is \cplus~\cite{giu04}, while it is answer set programming (ASP)~\cite{BrewkaEL16} for \hcpasp. In HCplan, outcomes of sensing actions are nondeterministically decided by external computations, whereas in \hcpasp\ they are nondeterministically decided by ASP. Moreover, \hcpasp\ keeps track of visited belief states and avoids recomputing subtrees of a hybrid conditional plan.

Although not compilation-based, the offline conditional planner \ascp\ is also related to \hcpasp: It is not hybrid, not parallel, and its input language is the action language ${\cal A}^c_{K}$; but it uses ASP to compute conditional plans. Based on 0-approximations~\cite{SonB01}, the idea is to approximate a belief state by a consistent set of fluent literals that are known to be true, and to define transitions considering what definitely holds and what may change at the next state. With such an approximation, the computational complexity of conditional planning becomes NP-complete. Given upper bounds for the height and the number of branches, \ascp\ generates a conditional plan with one call of an ASP solver based on an intelligent formulation of possible transitions. \hcpasp\ does not consider 0-approximations of belief states, but requires that the initial values of fluents that cannot be identified by any sensing action are known in advance, decided nondeterministically (e.g., by a disjunction over the possible values), or specified with respect to some assumptions (e.g., by defaults). Since \hcpasp\ is compilation-based, it calls an ASP solver multiple times to build a conditional plan.

There is a variety of recent work on hybrid planners that combine motion planning with classical planners. Some of them are based on modifications/introductions of search algorithms for motion/task planning~\cite{Asymov_book,Hauser2009,KaelblingL13,LagriffoulDBSK14,GaschlerPGRK13,SrivastavaFRCRA14}, some of them are based on formal methods~\cite{Plaku12,Dantam-RSS-16}, and some of them are based on modification of representation of domains~\cite{cal09,HertleDKN12,icra11,ErdemPS16}. \hcpasp\ is similar to the latter group, since feasibility checks are embedded in action descriptions via external atoms (in the spirit of semantic attachments in theorem proving~\cite{weyhrauch78}) without having to modify the classical planners or motion planners. We refer the reader to the recent studies~\cite{LagriffoulDBSK14,ErdemPS16} surveying and empirically analyzing some of these approaches.

%%%%%%%%%%%%%%%%%%%%%%%%%%%%%%%%%%%%%%%%%%%%%%%%%%%%%%%%%%%%

\section{Answer Set Programming}
\label{sec:asp}

Answer Set Programming (ASP)~\cite{BrewkaEL16}
is a form of knowledge representation and reasoning paradigm oriented towards
solving combinatorial search problems as well as knowledge-intensive
problems. The idea of ASP is to represent a problem as a ``program'' whose models (called
``answer sets''~\cite{gelfondL91})
correspond to the solutions. The answer sets for the given program
can be computed by special implemented systems called ASP
solvers, such as~\clingo~\cite{gekakasc14b}.

We consider ASP programs (i.e., nondisjunctive \hex\ programs~\cite{hex2005}) that consist of rules of the form
$$
\ii{Head} \lar A_1, \dots, A_m, \no\ A_{m+1}, \dots, \no\ A_n
$$
where $n \geq m \geq 0$, \ii{Head} is an atom or $\bot$, and each
$A_i$ is an atom or an external atom. A rule is called a
\textit{fact} if $m=n=0$ and a \textit{constraint} if \ii{Head} is
$\bot$. Please see Appendix~A for more details.

We model robotic action domains as ASP programs as described in~\cite{ErdemGL16}.
We use the ASP solver~\clingo with relevant feasibility checkers to compute hybrid plans (i.e., discrete task plans integrated with continuous feasibility checks).

There are several important aspects of ASP that are worth emphasizing here as they are closely relevant to our contributions in this paper.
First, it is possible to embed the results of external computations into ASP programs using special constructs, called ``external atoms'', in the spirit of semantic attachments in theorem proving~\cite{weyhrauch78}.
For instance, the ASP rule
$$
\ba l
\bot \lar \ii{at}(r,x_1,y_1,t), \ii{goto}(r,x_2,y_2,t), \no\ \ext{\ii{path\_exists}}{x_1,y_1,x_2,y_2}{}
\ea
$$
expresses that, at any step $t$ of the plan, a robot $r$
cannot move from $(x_1,y_1)$ to $(x_2,y_2)$ if it is not known that a
collision-free trajectory between them exists. Here, collision check is done
by the external atom $\ext{\ii{path\_exists}}{x_1,y_1,x_2,y_2}{}$ implemented in C++,
utilizing the bidirectional RRT algorithm~\cite{kuffnerrrt}. External atoms allow us to compute hybrid plans.

Second, it is possible to express defaults in ASP thanks to ``default negation'' $\no$.
The defaults allow us to model our assumptions about a dynamic robotic world with partial observability. For instance, consider a kitchen where a service robot sets up the table for a meal. Some locations are partially observable, but the table is fully observable (i.e., the robot knows what is on the table). The following ASP rule
$$%\beq
\ba l
\neg \ii{at}(o,\ii{Table},t)\lar \no\ \ii{at}(o,\ii{Table},t)
\ea
$$%\eeq{eq:default}
expresses that by default the objects $o$ in a kitchen are not assumed to be on the table unless they are known to be on the table.

Third, it is possible to express nondeterministic choice in ASP using ``choice expressions'' with ``cardinality constraints''. Choice expressions help us to model occurrences and non-occurrences of actions.
For instance, the following ASP rule
\beq
\{ \ii{sense}(\ii{at}(o),t) \}
\eeq{eq:sense-exogenous}
expresses that the action of sensing the location of an object can occur any time.
Choice expressions with cardinality constraints help us to model nondeterministic effects of sensing actions.  For instance, the following ASP rule
\beq
1\{ \ii{at}(o,l,t+1): loc(l) \}1 \lar \ii{sense}(\ii{at}(o),t)
\eeq{eq:sense-outcome}
describes that if sensing is applied to check the location of an object $o$ (i.e., $\ii{sense}(\ii{atObj}(o),t)$), then we know that the object $o$ is at one of the possible locations $l$; here, the location $l$ is nondeterministically chosen by the ASP solver.

Fourth, it is possible to express ``unknowns'' using ``cardinality expressions''; e.g., the rule
$$%\beq
\neg \ii{at}(o,m,t) \lar \{ \ii{at}(o,l,t):\ii{loc}(l)\}0
$$%\eeq{eq:unknown}
expresses that if object's location is not known (i.e., $\{ \ii{at}(o,l,t):\ii{loc}(l)\}0$) then it definitely can not be at a robot's hand $m$.

Fifth, we can express ``weak constraints'' to minimize, e.g., the number of sensing actions:
$$%\beq
\xleftarrow{\scriptstyle\sim}
\ii{senseAct}(t)\  [2@2,t]
$$%\eeq{eq:minsense}

Finally, the incremental computation of an answer set by an ASP solver, like \clingo, allows for minimization of makespans (i.e., lengths) of plans.

%%%%%%%%%%%%%%%%%%%%%%%%%%%%%%%%%%%%%%%%%%%%%%%%%%%%%%%%%%%%

\section{Hybrid Conditional Plans}

A hybrid conditional plan can be characterized as a labeled directed
tree $(V,E)$.
 The set $V = V_a \cup V_s$ of vertices consists of two
types of vertices. The vertices in $V_a$ (called actuation vertices) characterize hybrid
actuation actions (e.g., the robot's navigation and manipulation actions integrated with feasibility checks). The vertices in $V_s$
(called sensing vertices) characterize sensing actions or information gathering actions in
general (e.g., finding out or checking the location of an object). The leaves of the tree are in $V_a$.

The set $E$ of edges between vertices in $V$
characterizes the order of actions: an edge $(x,y)$ expresses that
the action denoted by the vertex $x$ is to-be executed before the
action denoted by~$y$. Each vertex in $V_a$ has at most one outgoing edge based on the
assumption that the actuation actions are deterministic. Each vertex
in $V_s$ has at least two outgoing edges. Each sensing action may
lead to different outcomes/observations.

Let us denote by $E_s$ the set of outgoing edges from
vertices in $V_s$. Then a labeling function maps every edge $(x,y)$
in $E_s$ by a possible outcome of the sensing action characterized
by~$x$.

%%%%%%%%%%%%%%%%%%%%%%%%%%%%%%%%%%%%%%%%%%%%%%%%%%%%%%%%%%%%

\section{A Parallel Algorithm for Computing a Hybrid Conditional Plan}

\hcpasp\ computes a hybrid conditional plan utilizing the expressive formalism and efficient solvers of ASP. \hcpasp\ first computes a hybrid sequential plan of sensing actions and actuation actions using ASP; this sequential plan characterizes a single branch of the hybrid conditional plan being constructed.  After identifying the sensing actions in this plan, for each sensing action and for each outcome of the sensing action, \hcpasp\ computes a hybrid conditional plan in parallel. Finally, it combines these hybrid conditional plans as children of the sensing nodes.
The overall algorithm is depicted in Algorithm~\ref{algMain}.
It is important to emphasize the novelties of our algorithm by explaining how it handles some of the challenges.

     \begin{algorithm}
        \small
        \caption{\hcpasp($\cal D$,$\cal P$,$S$,$nT$)}\label{algMain}
        \begin{algorithmic}[1]
            \INPUT{Robotic domain description $\cal D$, planning problem $\cal P$ with an initial belief state (i.e., the set of fluents with their known values) and goals, the set $S$ of all possible sensing actions where each sensing action $s$ is tupled with the set $O_s$ of all its possible outcomes, and the maximum number $nT$ of threads to run in parallel}
            \OUTPUT{A hybrid conditional plan, specified by its root $root$, if one exists; otherwise, failure.}

            \Statex // $hashT$ (initially empty) denotes a hash table: for each visited belief state $e$ with a key $e.id$, contains a hybrid conditional plan tree          
            \Statex // $taskQ$ (initially empty) denotes a queue of planning tasks (i.e., hybrid sequential planning problems) to be computed in parallel, where each task $\langle e,s,o_s\rangle$ is characterized by an initial belief state $e$, sensing action $s$ in $S$ to be executed at $e$ with an outcome $o_s$

            \Statex // create the first planning task, and push it to the task queue
            \State $e_0 \gets$ initial belief state according to $\cal P$
            \State $taskQ.Push(\langle e_0,null,null\rangle)$

            \Statex // while there are tasks in $taskQ$, solve them in parallel

            \While{$taskQ.Size()>0$}

              \ParFor{$\langle e,s,o_s\rangle\in taskQ$ \textbf{with} $nT$ \textbf{threads}}

                \Statex // extract the next planning task from $taskQ$
                \State $\langle e,s,o_s \rangle \gets taskQ.Pull()$

                \State ${\cal P} \gets$ copy and modify $\cal P$ with $\langle e,s,o_s\rangle$ so that the initial belief state is $e$ and, vif $s\neq\ii{null}$,  the first action of the computed plan is the sensing action $s$ in $S$ with outcome $o_s$ in $O_s$

                \State $existsP, P, H \gets$ compute a hybrid sequential plan $P$ for $\cal P$, with its history $H$, over $\cal D$

                \If{$existsP$}
                  \Statex // if $b_{\langle e,s,o_s\rangle}$ is the first branch computed
                  \If{$e == e_0$}
                     \State $b_{\langle e,s,o_s\rangle} \gets$ create a branch of the tree according to $H$, starting with the node $n_{\langle e,s\rangle}$ that characterizes the first action $s$ of $P$, and labeling the outgoing edges of sensing nodes in the branch accordingly
                     \State $root := b_{\langle e,s,o_s\rangle}$
                  \Else
                      \State $b_{\langle e,s,o_s\rangle} \gets$ create a branch of the tree according to $H$, starting with the node that characterizes the second action of $P$, and labeling the outgoing edges of sensing nodes in the branch accordingly
                  
                      \State $root \gets$ link the branch $b_{\langle e,s,o_s\rangle}$ to the existing hybrid conditional plan tree $root$ as a child of the sensing node $n_{\langle e,s\rangle}$ that characterizes $s$ and whose incoming edge describes $e$

                      \State $o_e \gets$ extract the observed outcome of $s$ executed at $e$, from $H$
                      \State $root \gets$ label the edge from $n_{\langle e,s\rangle}$ to $b_{\langle e,s,o_s\rangle}$ with $o_e$ in the tree $root$
                  \EndIf

                  \Statex //explore the branch $b_{\langle e,s,o_s\rangle}$ and expand it at sensing nodes considering other possible  outcomes

                  \For{$i=1\to |b_{\langle e,s,o_s\rangle}|-1$}
                      \State $e_i, a_i \gets$ identify from $H$, the belief state $e_i$ and the action $a_i$ executed at $e_i$
                       \Statex //check whether a subtree is already computed for $e_i$
                      \State $vs := hashT[e_i.id]$
                      \If{$vs$ is not null}
                            \State $n_{\langle e_i,a_i\rangle}\gets$ add $vs$ as a child of $n_{\langle e_i,a_i\rangle}$
                      \Statex // create a new planning task for other possible outcomes of $a_i$ and push it to the task queue
                      \ElsIf{$a_i$ is a sensing action in $S$}
                         \State $n_{\langle e_i,a_i\rangle}, o_{a_i} \gets$ identify the sensing node $n_{\langle e_,a_i\rangle}$ in branch $b_{\langle e,s,o_s\rangle}$ executed at $e_i$, and the label $o_{a_i}$ of its outgoing edge
                         \For{$o_s\in O_s$ different from $o_{a_i}$}
                             %\State $e'_i \gets$ update $e_i$ with $o_s
                             \State $taskQ.Push(\langle e_i,a_i,o_{s}\rangle)$
                         \EndFor
                      \EndIf
                     \State $hashT[e_i.id] \gets$ the hybrid conditional plan tree rooted at $n_{\langle e_i,a_i\rangle}$
                  \EndFor
                \EndIf
            \EndParFor
        \EndWhile
        \Return $root$
        \end{algorithmic}
\normalsize
    \end{algorithm}

\subsection{Computation of a single branch of the hybrid conditional plan} Considering the incomplete knowledge about the initial state, partial observability of the domain, deciding for (non)occurrences of sensing actions, nondeterministic effects of sensing actions, and necessity of feasibility checks for actuation and sensing actions for robotic applications, make it challenging to compute even a single branch of a hybrid conditional plan.
We deal with these challenges by describing the robotic domain and the planning problem in ASP.

\vspace{-2ex}
\paragraph{Representing actuation actions.} We formalize preconditions and effects of actuation actions in robotic domains as described in~\cite{ErdemGL16}. For some examples, please see Appendix~B.

\vspace{-2ex}
\paragraph{Integrating feasibility checks.} We integrate low-level feasibility checks into computation of a hybrid sequential plan, utilizing external atoms, as described in Section~\ref{sec:asp}, with respect to different methods of integration, as surveyed in \cite{ErdemPS16}.

\vspace{-2ex}
\paragraph{Representing belief states.} We assume that values $v$ of functional fluents $f(\bar{x})$ at step $t$ are represented by atoms of the form $f(\bar{x},\bar{v},t)$ in ASP.

We assume that some functional fluents are fully observable (i.e., known by the robot), while some are partially observable (i.e., can be identified by sensing actions).
If a functional fluent is fully observable then a uniqueness constraint
\beq
\lar 2 \{ f(\bar{x},\bar{v},t) : \ii{dom}_{\bar{x}}(\bar{v}\}
\eeq{eq:uniqueness}
and an existence constraint
$$%\beq
\lar \{ f(\bar{x},\bar{v},t) : \ii{dom}_{\bar{x}}(\bar{v})\}
$$%\eeq{eq:existence}
are included in the domain description.
If the fluent is partially observable then a uniqueness constraint~(\ref{eq:uniqueness}) is added only; the value of the fluent is identified when the relevant sensing action is performed.

For every functional fluent $f(\bar{x})$, if there exist some value $\bar{v}$ such that $f(\bar{x},\bar{v},t)$ belongs to an answer set then we believe that $f(\bar{x})$ is known at step $t$.  Otherwise, if there is no $\bar{v}$ such that $f(\bar{x})=\bar{v}$ belongs to an answer set for time step $t$, we believe that the value of $f(\bar{x})$ is not known. In that sense, partial assignments of values to fluents represent belief states in our framework.

\vspace{-2ex}
\paragraph{Representing initial values of fluents.} If the initial values of fully observable fluents are known, then they are specified by a set of facts. Otherwise, they are decided nondeterministically (e.g., by a disjunction over the possible values or by a choice rule):
$$
\{ f(\bar{x},\bar{v},0) \} \lar  \ii{dom}_{\bar{x}}(\bar{v})
$$
or specified with respect to some assumptions (e.g., by defaults).
$$
\neg f(\bar{x},\bar{v},0) \lar  \no\ f(\bar{x},\bar{v},0), \ii{dom}_{\bar{x}}(\bar{v}) .
$$

If the initial values of partially observable fluents are known, then they are specified by a set of facts.

\vspace{-2ex}
\paragraph{Representing nondeterministic hybrid sensing actions.}
We characterize sensing actions by atoms of the form $\ii{sense}(f(\bar{x}),t)$ which describes an occurrence of a sensing action that determines the value of a unique partially observable fluent $f(\bar{x})$ at time step $t$.

We suppose that sensing actions may occur exogenously at any time, and we represent this assumption by the following choice rule, like in rule~(\ref{eq:sense-exogenous}):
$$\{\ii{sense}(f(\bar{x}),t) \} .$$

We suppose that sensing actions $\ii{sense}(f(\bar{x}),t)$ are not executed if the value of the relevant fluent $f(\bar{x})$ is known.  We express this assumption by the following constraint:
$$\lar \ii{sense}(f(\bar{x}),t), 1\{ f(\bar{x},\bar{v},t) : \ii{dom}_{\bar{x}}(\bar{v})\} .$$

After that, we describe that whenever a sensing action $\ii{sense}(f(\bar{x}),t)$ is executed, the observed outcome $\bar{v}$ is nondeterministically chosen among possible values of $f(\bar{x})$. We express this nondeterministic choice of an outcome using choice expressions and cardinality constraints in ASP, like in rule~(\ref{eq:sense-outcome}):
$$
1\{ f(\bar{x},\bar{v},t+1) : \ii{dom}_{\bar{x}}(\bar{v})\} 1 \lar \ii{sense}(f(\bar{x}),t) .
$$

Sometimes, executability of a sensing action necessitates further conditions to hold.
For instance, to check whether a plate is clean or not, the robot has to be holding the plate.
The preconditions of a sensing action may involve feasibility checks as well. For instance, checking what is on the table is possible if the camera is ``reachable'' to a position over the table. This can be expressed by a constraint as follows:
$$
\lar \ii{sense}(\ii{at}(o),t), \no\ \ext{\ii{reachable}}{Table}{} .
$$

\vspace{-2ex}
\paragraph{Incremental computation of a hybrid sequential plan of sensing actions and actuation actions.} Given a domain description and a planning problem as described above, the ASP solver \clingo\ can compute a plan to reach a goal state from a possibly incomplete initial state. \clingo\ first grounds this whole program and then computes an answer set for the ground program; the answer set characterizes the plan.

We improve this computation (grounding and answer set finding), by ensuring that \clingo\ incrementally performs this computation by increasing the upper bound on the maximum time step one by one, but without having to ground the whole theory every time. This is possible thanks to the ``incremental mode'' of \clingo.

\subsection{Parallel computation of subtrees} Our algorithm maintains a queue $taskQ$ of tasks~$\langle e,s,o_s\rangle$ where $e$ is an initial belief state $e$, $s$ is a sensing action, $o_s$ is an outcome of $s$. Initially, $taskQ$ contains the task $\langle e_0,\ii{null},\ii{null}\rangle$ only (Algorithm~\ref{algMain}, Line~2).  Each task~$\langle e,s,o_s\rangle$ in $taskQ$ describes a hybrid sequential planning problem with the initial state $e$ and the goal as same as in the hybrid conditional planning problem~${\cal P}$ (Lines~5\&6). This planning problem has further constraints, if $s$ and $o_s$ are not \ii{null}: $s$ is a sensing action executed at belief state $e$ and the outcome of this execution is~$o_s$ (Line~7). Note that these constraints can be easily expressed in ASP.

For every task $\langle e,s,o_s\rangle$, an ASP solver computes a hybrid sequential plan~$P=\langle a_0,a_1,...,a_k\rangle$ and its history~$H=\langle b_0,a_0,b_1,a_1,...,b_k,a_k,b_{k+1}\rangle$, where $s{=}a_0$, $e=b_0$, each $a_i$ is an actuation action or a sensing action executed at a belief state $b_i$ and reaching a belief state $b_{i+1}$ ($0\leq i<k$), the last action $a_k$ is an actuation action, and the last belief state $b_{k+1}$ is a goal state according to~${\cal P}$. Such a plan~$P$ describes a branch $b_{\langle e,s,o_s\rangle} = \langle n_{\langle b_0,a_0\rangle}, n_{\langle b_1,a_1\rangle}, ..., n_{\langle b_k,a_k\rangle} \rangle$ of the tree that starts at the node $n_{\langle e,s\rangle} = n_{\langle b_0,a_0\rangle}$ and ends at the leaf $n_{\langle a_k,b_k\rangle}$. For every sensing action $a_i$ in $P$, the edge from $n_{\langle b_i,a_i\rangle}$ to $n_{\langle b_{i+1},a_{i+1}\rangle}$ is labeled by its outcome $o_{a_i}$ observed in $b_{i+1}$. Once we compute all the branches that start at a sensing node, we can combine them at that node (Lines~9--16).

The tasks are added to the queue $taskQ$ at different times: for every other outcome of every sensing node in a branch $b_{\langle e,s,o_s\rangle}$, a task is constructed (Lines~17--26). Since the tasks in $taskQ$ are not dependent on each other, they are solved in parallel utilizing at most $nT$ threads (Line~4).

{\em Discussion about correctness.} Every branch of the tree from the root to a leaf constructed incrementally as described above characterizes a hybrid sequential plan, under the assumptions that a plan can be constructed by merging two plans: $\cal D$ respects the Markov principle (so actions do not have delayed effects), and $\cal D$ does not contain any temporal constraints that ensure/prevent occurrences of actions or observations of fluents in some order, or any constraints to guarantee some conditions over the whole hybrid sequential plan (e.g., a constraint on the total cost of actions).
Indeed, let $P=\langle a_0,a_1,...,a_k\rangle$ be a hybrid sequential plan computed for a planning problem $\cal P$ with respect to a domain description $\cal D$, where each $a_i$ is an actuation action or a sensing action ($0\leq i<k$) and the last action $a_k$ is an actuation action. Let $H=\langle b_0,a_0,b_1,a_1,...,b_k,a_k,b_{k+1}\rangle$ be a history of this plan,  where every action $a_i$ is executed at a belief state $b_i$ and reaching a belief state $b_{i+1}$ ($0\leq i<k$), and the last belief state $b_{k+1}$ is a goal state according to~${\cal P}$. For every sensing action $a_j$ in $P$ ($0< j < k$) with an outcome $o_{a_j}$ observed at $b_{j+1}$, a task $\langle b_j,a_j,o_{a_j}\rangle$ is constructed and pushed into $taskQ$. Consider one of these tasks: $\langle b_j,a_j,o_{a_j}\rangle$. Let $P'=\langle a_j=a'_0,a'_1,...,a'_{k'}\rangle$ be a hybrid sequential plan, with its history $H'=\langle b_j=b'_0,a'_0,b'_1,a'_1,...,b'_{k'},a'_{k'},b'_{k'+1}\rangle$, computed for the planning problem characterized by this task. Then, due to the assumptions mentioned above, the sequence $\langle a_0,a_1,...,a_j,a'_1,...,a'_{k'} \rangle$ of actions is also a hybrid sequential plan computed for a planning problem $\cal P$, with a history $\langle b_0,a_0,b_1,a_1,...,b_j,a_j,b'_1,a'_1,...,b'_{k'},a'_{k'},b'_{k'+1}\rangle$: the first part of the plan, $\langle a_0,a_1,...,a_j \rangle$, does not prevent the last part of the plan, $\langle a'_1,...,a'_{k'} \rangle$, and vice versa.
Note that these assumptions can be waived if we modify our algorithm slightly by computing every branch of the tree from the root, instead of constructing branches by combining sub-branches: every task $\langle b_j,a_j,o_{a_j}\rangle$ is understood as a planning problem obtained from ${\cal P}$ by adding constraints to ensure that $\langle b_0,a_0,b_1,a_1,...,b_j,a_j\rangle$ is a prefix of the history of the plan and that $o_{a_j}$ is observed right after execution of $a_j$. On the other hand, note that the hybrid sequential plans computed from the root have larger makespans (and thus they lead to computations over larger ASP programs) compared to the plans computed from intermediate states.

\subsection{Avoiding recomputation of subtrees} Our algorithm keeps a hash table $hashT$ that maps a belief state to a hybrid conditional plan computed from that state (Lines~18--21,26). In this way, identical parts of a hybrid conditional plan are not computed repeatedly. This then allows the representation of a hybrid conditional plan more compactly as a directed acyclic graph (DAG).

For some domains, when the desired goal relies on a subset of fluents only, we define ``equivalence classes'' of belief states according to these fluents. This allows us to utilize the hash table with respect to these equivalence classes. If belief states $e$ and $e'$ belong to the same equivalence class, and if a hybrid conditional plan is computed for $e$, then we utilize this tree also for $e'$ despite their differences.

For instance, consider the \emph{colorball} planning domain~\cite{AlborePG09} used as a benchmark for conditional planners. In this domain,
the goal is to dispose a given set of balls with unknown locations and colors into thrash boxes according to their colors. As long as a ball is disposed into the trash, its color or location does not affect the rest of the plan anymore. So we can safely say that two different belief states are equivalent if they only differ in color and/or location of any ball that is trashed.
Fortunately, we can specify this information in ASP (in a separate file) by identifying what parts of belief states could be ignored as follows:
$$
\ba l
\ii{redundant}(\ii{color}(b,c,t)) \lar \ii{trashed}(b,t) \\
\ii{redundant}(\ii{location}(b,l,t)) \lar \ii{trashed}(b,t) .
\ea
$$
When a ball $b$ is thrashed, its color $c$ and location $l$ are irrelevant to belief state comparisons.
Here $\ii{redundant}$ is a reserved predicate for our planner to identify partial belief states for comparisons.

\subsection{Advantages of hybridity of conditional plans}

\begin{figure}[bth]
      \vspace{-0.5\baselineskip}
\begin{tabular}{cc}
    \includegraphics[scale=0.22]{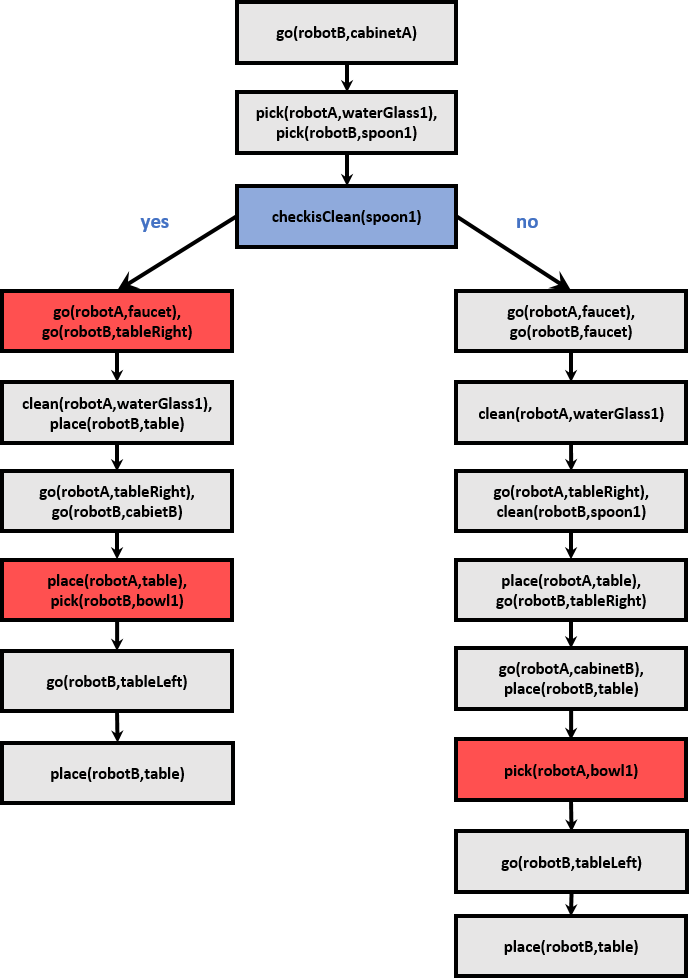}
    &
    \includegraphics[scale=0.22]{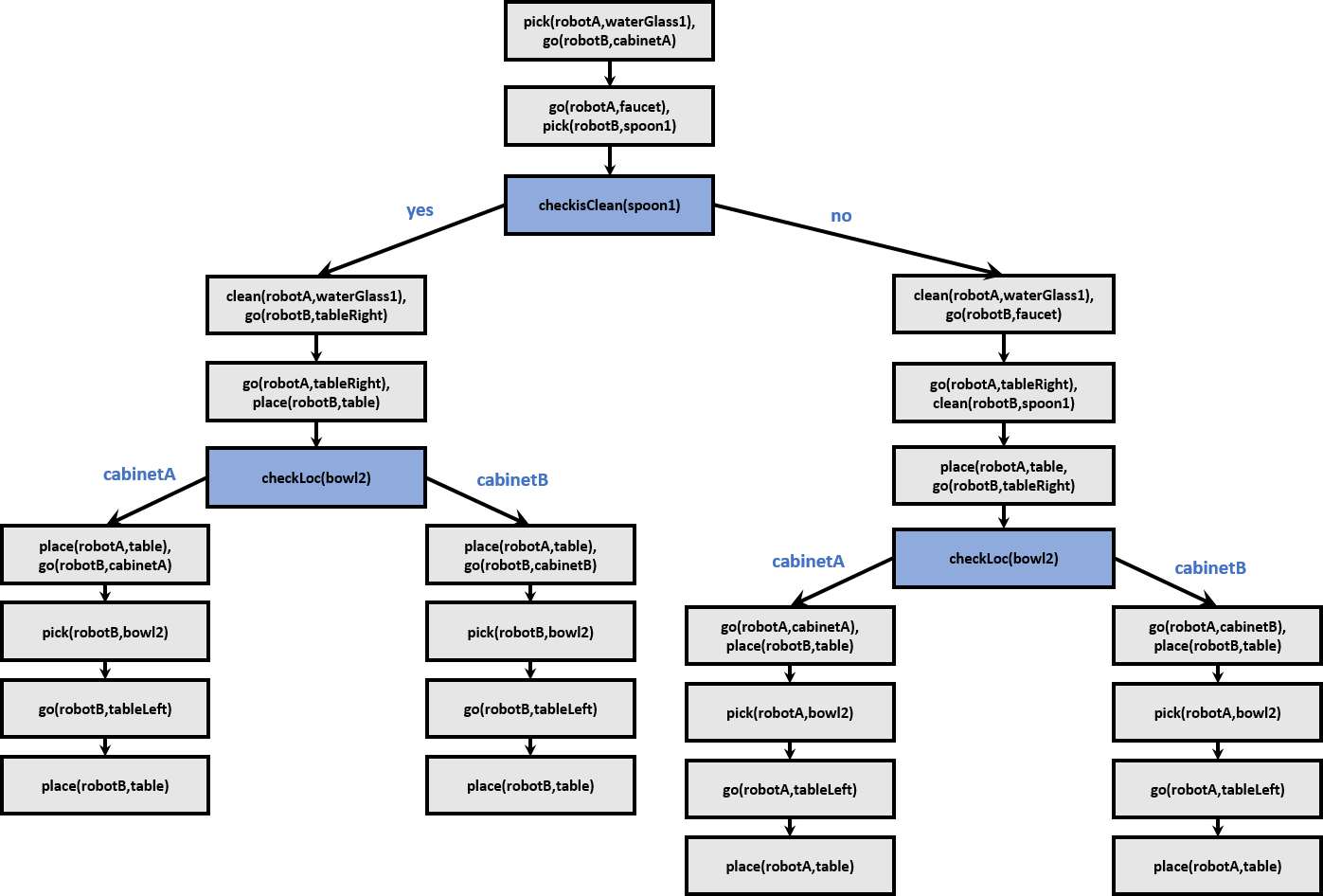} \\
    (a) & (b)
\end{tabular}
\vspace{-0.5\baselineskip}
    \caption{Two hybrid conditional plans generated (a) without and (b) with feasibility checks.}
    \label{fig:plans}
      \vspace{-.5\baselineskip}
\end{figure}

Since each branch of a conditional plan depicts a possible execution of actuation/sensing actions to reach a goal, it is essential that these actions are checked against relevant feasibility constraints in real-world applications.  In robotics applications, these constraints are required for collision-free navigation and reachable/graspable manipulation, as depicted in Figure~\ref{fig:plans} with two conditional plans computed for a robotics scenario, where two bimanual mobile manipulators are responsible for setting up a kitchen table: none of the branches of the non-hybrid plan (Figure~\ref{fig:plans}(a)) is executable in real world, since the actuation actions (denoted red) are not feasible; whereas every branch of the hybrid plan  (Figure~\ref{fig:plans}(b)) is feasible in real world. See Appendix~C, for more details about the domain.

%%%%%%%%%%%%%%%%%%%%%%%%%%%%%%%%%%%%%%%%%%%%%%%%%%%%%%%%%%%%%

\section{Experimental Evaluations}

\paragraph{Comparisons with the hybrid compilation-based conditional planner HCPlan}

We have compared \hcpasp\ (with \clingo~4.5.4) with HCPlan (with \ccalc\ using \minisat~2.2.0) over the robotics kitchen \emph{table setting} domain of \cite{NoumanYEP16}.
This domain models a dynamic service robotics scenario, where a bimanual mobile manipulator is responsible for setting up a kitchen table by navigating around a kitchen to manipulate kitchenware. The robot can perform sensing actions to identify the locations of objects, cleanliness of objects, and to determine the type of food requested by the user. The goal is to for the robot to setup the table for lunch, with proper kitchenware for the requested food type. This domain includes several low-level feasibility checks, such as kinematic reachability, bimanual motion planning for robot arms, and collision-free navigation planning.
For our experiments, we have constructed 12 instances of this domain, with solutions whose heights range over 16--33 and branching factor ranges over 2--4.

All experiments are conducted on a PC workstation running Ubuntu~16.04 on 16~2.4GHz Intel E5-2665 CPU cores with 64GB memory and an SSD hard drive. As suggested by the PRE integration method of \cite{ErdemPS16}, the feasibility checks are performed in advance as in \cite{NoumanYEP16} and cached in a hash table, for both planners. A 5~hour threshold is used to compute plans, and computations that exceed this threshold are marked as Time Out (TO). For parallel implementations of \hcpasp\ and HCPlan, the maximum number of available threads are limited to~20. \hcpasp\ generates a more compact representation of the tree as a DAG due to re-use of subtrees whereas HCplan generates a tree. Therefore, we report the tree sizes of plans.

Table~\ref{tab:comparisonPlanners} summarizes some results of our experiments. For each instance and the computed hybrid conditional plan, we present the maximum branch length (i.e., makespan of the longest hybrid sequential plan), the number of sensing nodes in the tree (i.e., total number of sensing actions in the plan), the number of leaves of the tree (i.e., number of different hybrid sequential plans represented in the tree), the number of nodes in the tree (i.e., total number of actuation and sensing actions in the plan), the computation time in seconds (i.e., walltimes for the threads with longest computations). The instances are arranged in ascending order of tree sizes.

\begin{table}[b]
\vspace{-1.5\baselineskip}
    \caption{Comparison of \hcpasp\ with HCPlan on the table setting domain.}
    \label{tab:comparisonPlanners}
    \vspace{-.5\baselineskip}
\setlength{\tabcolsep}{0.5pt}
    \begin{center}
        {
            %\resizebox{\columnwidth}{!}{
                \begin{tabular}{|c||r|r|r|r|r|r|r|r|r|r|}
					\cline{1-11}
					\multicolumn{1}{|c||}{\multirow{2}{*}{Scen.}} & \multicolumn{2}{c|}{Max Branch Length} & \multicolumn{2}{c|}{No. Sensing Nodes} & \multicolumn{2}{c|}{No. of Leaves} & \multicolumn{2}{c|}{Tree Size} & \multicolumn{2}{c|}{Time} \\ \cline{2-11}
					\multicolumn{1}{|c||}{} & \multicolumn{1}{l|}{HCplan} & \multicolumn{1}{l|}{\hcpasp}& \multicolumn{1}{l|}{HCplan} & \multicolumn{1}{l|}{\hcpasp} & \multicolumn{1}{l|}{HCplan} & \multicolumn{1}{l|}{\hcpasp} & \multicolumn{1}{l|}{HCplan} & \multicolumn{1}{l|}{\hcpasp} & \multicolumn{1}{l|}{HCplan} & \multicolumn{1}{l|}{\hcpasp}\\ \cline{1-11}
					%1 & 16 & 16 & 1 & 1 & 3 & 3 & 37 & 40 & 11.27 & 11.09  \\ \cline{1-11}
					%2 & 16 & 16 & 2 & 1 & 4 & 3 & 42 & 40 & 12.21 & 10.95  \\ \cline{1-11}
					1 & 17 & 17 & 4 & 3 & 12 & 11 & 147 & 114 & 19.66 & 18.85  \\ \cline{1-11}
					2 & 21 & 21 & 11 & 9 & 19 & 17 & 168 & 186 & 25.91 & 47.70  \\ \cline{1-11}
					3 & 21 & 20 & 15 & 8 & 35 & 25 & 306 & 243 & 31.38 & 26.79  \\ \cline{1-11}
					4 & 23 & 23 & 20 & 24 & 31 & 32 & 294 & 283 & 35.24 & 106.77  \\ \cline{1-11}
					5 & 23 & 22 & 31 & 24 & 45 & 32 & 398 & 329 & 41.66 & 180.80  \\ \cline{1-11}
					6 & 21 & 25 & 31 & 26 & 72 & 40 & 566 & 377 & 33.8 & 53.82  \\ \cline{1-11}
					7 & 26 & 26 & 49 & 54 & 102 & 68 & 771 & 615 & 45.73 & 164.48  \\ \cline{1-11}
					8 & 26 & 26 & 53 & 54 & 91 & 68 & 676 & 615 & 38.96 & 164.11  \\ \cline{1-11}
					9 & 29 & 24 & 103 & 58 & 222 & 96 & 1535 & 698 & 53.88 & 239.26  \\ \cline{1-11}
					10 & 29 & 31 & 231 & 224 & 347 & 280 & 2296 & 1844 & 67.99 & 202.25  \\ \cline{1-11}
					11 & 29 & 29 & 227 & 156 & 349 & 218 & 2524 & 1591 & 76.28 & 280.52  \\ \cline{1-11}
					12 & 33 & 31 & 1483 & 576 & 2178 & 773 & 16119 & 5164 & 390.29 & 973.28  \\ \cline{1-11}
				\end{tabular}
			}
	\end{center}
\end{table}

HCplan uses \ccalc\ with incremental grounding. Also, \hcpasp\ uses \clingo\ with incremental grounding. Therefore, both planners try to minimize the makespans of hybrid sequential plans. Although the planners start with a shortest hybrid sequential plan of the same makespan, due to the occurrences of different actions in the plans in different orders, the trees may have different heights in the end. This can be observed from the maximum branch lengths.

Using weak constraints, \hcpasp\ further tries to minimize the number of sensing actions, once it minimizes actuation actions. Due to this, the trees computed by \hcpasp\ have less number of sensing actions, and often with less number of leaves and nodes. Depending how the tree changes, the computation time may increase due to such further optimizations.

Due to parallel computations of hybrid sequential plans and subtrees, the computation times reflect in general the computation time of the longest hybrid sequential plan. Indeed, for the longest branches of the trees of Instance 12, \clingo\ takes much longer time than \ccalc.

\vspace{-2ex}
\paragraph{Comparisons with non-hybrid compilation-based conditional planners}

We have also compared \hcpasp\ with non-hybrid compilation-based planners CLG, DNFct and PO-PRP, over two standard benchmarks~\cite{AlborePG07,AlborePG09}. In the \emph{colorball} domain, the goal is to dispose a given set of $x$ balls with unknown locations and colors into thrash boxes according to their colors; the environment is represented as a $n\times n$ grid. The location and color of a ball can be observed when the agent and the ball are in the same cell. In the \emph{doors} domain, the goal is to reach a goal location by moving in a $n\times n$ grid with hidden doors; open doors can be detected by sensing actions. We have formalized these domains in ASP, by allowing multi-valued sensing outcomes and concurrency of actuation actions. Since CLG generates a tree, while DNFct and PO-PRP generate DAGs, both DAG sizes and tree sizes are reported for \hcpasp.

Table~\ref{tab:cplanners} presents the results of these experiments. CB $x-n$ (resp. Doors~$n$) denotes instances of the \emph{colorball} (resp. \emph{doors}) domain. For each instance, the size of DAG and/or tree (the number of nodes), the maximum branch length, and total computation time in seconds are reported in the table. MO indicates the planner did not have enough memory to complete the problem. Note that the available version of PO-PRP has a 5GB memory limit.

\begin{table}[b]
\centering
\vspace{-1.5\baselineskip}
\setlength{\tabcolsep}{0.5pt}
\caption{Comparison of offline compilation-based conditional planners.}
\vspace{-.25\baselineskip}
\label{tab:cplanners}
\begin{tabular}{|l||r|r|r||r|r|r||r|r|r||r|r|r|r|}
\cline{1-14}
\multicolumn{1}{|c||}{}          & \multicolumn{3}{c||}{CLG} & \multicolumn{3}{c||}{DNFct} & \multicolumn{3}{c||}{PO-PRP}    & \multicolumn{4}{c|}{\hcpasp} \\
\cline{1-14}
\multicolumn{1}{|l||}{}          & \multicolumn{1}{c|}{Tree}  & \multicolumn{1}{c|}{Max} & \multicolumn{1}{c||}{Time} & \multicolumn{1}{c|}{DAG}  & \multicolumn{1}{c|}{Max} & \multicolumn{1}{c||}{Time} & \multicolumn{1}{c|}{DAG}  & \multicolumn{1}{c|}{Max} & \multicolumn{1}{c||}{Time} & \multicolumn{1}{c|}{Tree}  & \multicolumn{1}{c|}{DAG}  & \multicolumn{1}{c|}{Max} & \multicolumn{1}{c|}{Time} \\
\multicolumn{1}{|l||}{}          & \multicolumn{1}{c|}{Size}  & \multicolumn{1}{c|}{Depth } & \multicolumn{1}{c||}{[sec]} & \multicolumn{1}{c|}{Size}  & \multicolumn{1}{c|}{Depth} & \multicolumn{1}{c||}{[sec]} & \multicolumn{1}{c|}{Size}  & \multicolumn{1}{c|}{Depth} & \multicolumn{1}{c||}{[sec]} & \multicolumn{1}{c|}{Size}  & \multicolumn{1}{c|}{Size}  & \multicolumn{1}{c|}{Depth} & \multicolumn{1}{c|}{[sec]} \\
\cline{1-14}
CB 3-1 		& 95		& 21	& 0.04			& 35		& 19	& 0.03			&89		&23		&0.02		&81  & 51			&17		&0.7	\\
CB 3-2 		& 2641		& 34	& 0.63			& 609		& 40	& 0.47			&1655		&48		&0.16		&1800 & 415		&28		&2	\\
CB 3-3 		& 60924		& 48	& 17.86			& 8021	& 57	& 8.45			&33247	&73		&3.96		&38977 &3395	&41		&29.8	\\
CB 3-4 		& 1329235	& 61	& 542			& 93379	& 72	& 268			&MO			&MO		&MO			&805768 &32593	&48		&928	\\
\cline{1-14}
CB 4-1 		& 295		& 52	& 0.2			& 67		& 36	& 0.04			&271		&46		&0.06		&242 & 100		&33		&1.4	\\
CB 4-2 		& 20050		& 74	& 11.75			& 1928	& 70	& 2.85			&12432	&90		&1.46		&14067 &1362	&58		&9.9	\\
CB 4-3 		& 1136920	& 100	& 966			& 39678	& 100	& 227			&MO			&MO		&MO			&708580 &27942	&84		&681\\
\cline{1-14}
CB 5-1 		& 586		& 65	& 0.59			& 117		& 68	& 0.95			&556		&78		&0.1		&524 &172		&63		&4	\\
CB 5-2 		& 72817		& 107	& 100			& 5153	& 113	& 18			&49767	&169		&15.22		&53576 &3479	&98		&38.5	\\
CB 5-3 		& TO		& TO	& TO			& 134008	& 157	& 2338			&MO			&MO		&MO			&4718549 &13879 &149	&13173	\\
\cline{1-14}
CB 9-1 		& 3385		& 197	& 44.45			& 351		& 184	& 47.28			&2607		&279		&1.1		&3210 & 621		&211	&77.5	\\
CB 9-2 		& 1700499	& 312	& 4918			& 43554	& 360	& 1564			&MO			&MO		&MO			&1238013 &42213	&378	&6779	\\
\cline{1-14}
\multicolumn{1}{|l||}{Doors 5}           & 144		& 24	& 0.18			& 146		& 26	& 0.05			&105		&28		&0.06		&156 & 68		&27		&1.2	\\
\multicolumn{1}{|l||}{Doors 7}           & 2153		& 51	& 3.38			& 2193	& 53	& 0.79			&1282		&60		&0.34		&2381 &179		&59		&6.5	\\
\multicolumn{1}{|l||}{Doors 9}           & 46024		& 95	& 188		& 44998	& 89	& 29.76		&23897	&104		&7.92		&52065 &381		&109	&47.4	\\
\multicolumn{1}{|l||}{Doors 11}          & 1213759	& 135	& 19228			& 1161651	& 124	& 1239			&MO			&MO		&MO			&1242179 &776	&167	&265	\\
\cline{1-14}
\end{tabular}
\vspace{-.5\baselineskip}
\end{table}

The results indicate that, for most of the scenarios, \hcpasp\ generates shorter trees and smaller DAGs.
\hcpasp\ scales in time better then CLG and PO-PRP, with a performance very close to that of DNFct. While PO-PRP shows remarkable time performance in available instances, the time and memory scaling of this planner seem poorer than others. \hcpasp\ outperforms other planners in generated DAG size which has high correlation with time and memory efficiency. The DAGs computed by \hcpasp\ are much smaller than the trees they represent. This shows the usefulness of identifying equivalence classes to detect similar states.

\vspace{-2ex}
\paragraph{Comparisons with \ascp}
As discussed in Section~\ref{sec:related}, although not compilation-based, the offline non-hybrid conditional planner \ascp\ also uses ASP to compute conditional plans. Therefore, we have compared these two ASP-based conditional planners, over one of the benchmarks of \ascp: Bomb in the Toilet with Sensing Actions (BTS)~\cite{WeldAS98}. In our experiments with \ascp, we have directly used the ASP encoding of BTS transformed from ${\cal A}^c_K$ with the ASP solver \clingo. According to the results (presented in Appendix~D), finding a tree with one call of \clingo\ (using \ascp) takes more time, compared to computing and combining the branches of the tree in parallel (using \hcpasp). For instance, for $m=17$, it takes more than an hour to compute a tree with \ascp\, whereas it takes about a second for \hcpasp.

\vspace{-2ex}
\paragraph{Evaluations of different versions of \hcpasp}

Using the \emph{colorball} and \emph{doors} domains, we also experimentally evaluate the following three versions of \hcpasp:
\textbf{${v_0}$}  (base planner with parallelism, revisited states, and incremental planning),
\textbf{${v_1}$} (${v_0}$  improved with equivalence classes),
\textbf{${v_2}$} (${v_1}$  improved with more efficient representation of the domain to capture multi-valued sensing outcomes and concurrency of actuation actions). The version ${v_2}$ is used in Tables~\ref{tab:comparisonPlanners} and~\ref{tab:cplanners}.

Table~\ref{tab:versions} presents the experimental results. For each instance, the size of DAG, the maximum branch length, parallel efficiency (i.e., ratio of sequential computation time to (20 $\times$ parallel computation time)), and the total computation time in seconds are reported  in the table.

\begin{table}[b]
\centering
\vspace{-1.5\baselineskip}
\setlength{\tabcolsep}{0.5pt}
\caption{Comparison of different versions of \hcpasp}
\vspace{-.25\baselineskip}
\label{tab:versions}
{
{
\begin{tabular}{|l||r|r|r|r||r|r|r|r||r|r|r|r|}
\cline{1-13}
\multicolumn{1}{|c||}{} & \multicolumn{4}{c||}{${v_0}$} & \multicolumn{4}{c||}{${v_1}$} & \multicolumn{4}{c|}{${v_2}$} \\
\cline{1-13}
%\hline
 & DAG & Max & Eff. & Time & DAG & Max & Eff. & Time & DAG & Max & Eff. & Time  \\
 & Size & Depth & \% & sec & Size & Depth & \% & sec & Size & Depth & \% & sec  \\
\cline{1-13}
CB 3-1      & 63            & 20        & 10        & 0.8           & 63            & 20        & 10        & 1.1           & 51            & 17        & 11        & 0.7       \\
CB 3-2      & 728           & 39        & 60        & 3.3           & 635           & 39        & 58        & 3.6           & 415           & 28        & 78        & 2     \\
CB 3-3      & 6960      & 52        & 40        & 93            & 4780      & 52        & 34        & 90            & 3395      & 41        & 97        & 29.8      \\
CB 3-4      & 113466    & 67        & 89        & 6738          & 55904     & 69        & 26        & 5835          & 32593     & 48        & 100       & 928       \\
\cline{1-13}
CB 4-1      & 117           & 37        & 11        & 1.6           & 121           & 40        & 12        & 1.8           & 100           & 33        & 12        & 1.4       \\
CB 4-2      & 1934      & 66        & 64        & 12.7          & 1503      & 69        & 57        & 12.8          & 1362      & 58        & 93        & 9.9       \\
CB 4-3      & 34074     & 101       & 98        & 788           & 24500 & 102       & 84        & 635           & 27942     & 84        & 99.8      & 681       \\
\cline{1-13}
CB 5-1      & 181           & 68        & 8         & 4.6           & 181           & 68        & 8         & 5.5           & 172           & 63        & 10        & 4     \\
CB 5-2      & 4939      & 121       & 65        & 65.5          & 3872      & 118       & 49        & 76.6          & 3479      & 98        & 96        & 38.5      \\
CB 5-3      & 123623    & 166       & 99        & 10023         & 109157    & 162       & 98        & 8168          & 138793   & 149       & 100       & 13173     \\
\cline{1-13}
CB 9-1      & 634           & 220       & 8         & 102           & 634           & 220       & 8         & 106           & 621           & 211       & 10        & 77.5      \\
CB 9-2      & 45584 & 397       & 96        & 8916          & 43235 & 397       & 96        & 9504          & 42213 & 378       & 98        & 6779      \\
\cline{1-13}
Doors 5           & 158           & 27        & 13        & 1             & 68            & 27        & 10        & 1.2 & \multicolumn{4}{c|}{}\\
Doors 7           & 2410      & 65        & 54        & 7.6           & 179           & 59        & 13    &6.5  &\multicolumn{4}{r|}{}         \\
Doors 9           & 49381     & 109       & 99        & 523           & 381       & 109       & 19        & 47.4 &\multicolumn{4}{r|}{}   \\
Doors 11          & TO                & TO        & TO        & TO            & 776       & 167       & 30        & 265&\multicolumn{4}{r|}{}\\
\cline{1-13}
\end{tabular}}}
\vspace{-.5\baselineskip}
\end{table}

The \emph{colorball} domain well demonstrates the effects of improvements in different versions of \hcpasp, while the \emph{doors} domain is useful to evaluate the potential of partial state comparisons, where a problem can be divided into sub-problems. Since the \emph{doors} domain is very simple, it does not allow for ${v_2}$, as there is no sensing outcome that can converted to multi-valued format and there is only a single actuation action not allowing for concurrency. 

Comparison of computation times of ${v_0}$ and efficiency values indicate that parallelism reaches its full potential with up to 99.9\% efficiency as the problem size gets larger. This is expected as the number of tasks that can be computed at any time depends on the number of sensing actions and the outcomes at each branch.
From the results of ${v_1}$, one can observe that utilization of equivalence classes result in highly reduced DAG size and number of branches explored, significantly improving the computation time.

From the results of ${v_2}$, one can observe that the use of multi-valued functional fluents and thus multi-valued sensing outcomes (e.g., identifying the color of a ball) decrease the length of branches compared to the case with boolean fluents and thus sensing outcomes (e.g., identifying that a ball has a specified color), since outcome of a single multi-valued sensing action can be obtained by an exponential number of boolean sensing actions.
Similarly, the use of concurrent actions reduce the length of plans, since multiple actions can be executed at the same time.

\section{Conclusion}

We have introduced an offline parallel compilation-based hybrid conditional planner, called \hcpasp, utilizing various features and utilities of ASP solver \clingo. From the representation point of view, compared to many existing offline compilation-based conditional planners, \hcpasp\ allows multi-valued functional fluents, multi-valued sensing actions, explicit representation of occurrences, preconditions and nondeterministic effects of sensing actions, weak constraints to minimize sensing actions or actuation actions, concurrency of actuation actions, and integration of feasibility checks into action descriptions. From the computation point of view, \hcpasp\ re-uses  earlier computations due to partial state comparisons via equivalence classes specified in ASP, generates trees in a more compact form of DAGs, performs incremental computations of branches (i.e., hybrid sequential plans), and utilizes parallel computation of branches in a subtree as well as parallel computation of subtrees. Due to these novelties, \hcpasp\ not only provides a new conditional planner with a more expressive representation language and promising computational performance, but also provides a useful tool for hybrid robotic applications with incomplete knowledge and partial observability.
Our planner \hcpasp\ and the benchmarks are available at {\small\url{http://cogrobo.sabanciuniv.edu/?p=1073}}.

\smallskip\noindent{\bf Acknowledgments.} This work is partially supported by the Science Academy (BAGEP). We thank T.~C. Son and H. Geffner for useful discussions about \hcpasp, and A. Albore, T.~C. Son and C. Muise for their helps with CLG, DNFct, PO-PRP respectively.

%%%%%%%%%%%%%%%%%%%%%%%%%%%%%%%%%%%%%%%%%%%%%%%%%%%%%%%%%%%%
\begin{appendix}

%%%%%%%%%%%%%%%%%%%%%%%%%%%%%%%%%%%%%%%%%%%%%%%%%%%%%%%%%%%%%%%%%%%%%%%%%
\section{ASP Programs} 

%Answer Set Programming (ASP)~\cite{BrewkaEL16}
%is a form of knowledge representation and reasoning paradigm oriented towards
%solving combinatorial search problems as well as knowledge-intensive
%problems.
%The idea of ASP is to represent a problem as a ``program'' whose models (called
%``answer sets''~\cite{gelfondL91})
%correspond to the solutions. The answer sets for the given program
%can be computed by special implemented systems called ASP
%solvers, such as~\clingo~\cite{gekakasc14b}.

We consider ASP programs (i.e., nondisjunctive \hex\ programs~\cite{hex2005})
that consist of rules of the form
$$
\ii{Head} \lar A_1, \dots, A_m, \no\ A_{m+1}, \dots, \no\ A_n
$$
where $n \geq m \geq 0$, \ii{Head} is an atom or $\bot$, and each
$A_i$ is an atom or an external atom. HEX programs can be extended by allowing
classical negation $\neg$ in front of atoms.
A rule is called a \textit{fact} if $m=n=0$ and a \textit{constraint} if \ii{Head} is
$\bot$.

An external atom is an expression of the form
$\ext{g}{y_1,\dots,y_k}{x_1,\dots,x_l}$ where $y_1,\dots,y_k$ and
$x_1,\dots,x_l$ are two lists of terms (called input and output
lists, respectively), and $\&g$ is an external predicate name.
Intuitively, an external atom provides a way for deciding the truth
value of an output tuple depending on the extension of a set of
input predicates. External atoms allow us to embed results of
external computations into ASP programs. They are usually
implemented in a programming language of the user's choice, like
C++. For instance, the following rule
$$\ba l
\bot \lar \ii{at}(r,x_1,y_1,t), \ii{goto}(r,x_2,y_2,t),\\
\qquad \no\ \ext{\ii{path\_exists}}{x_1,y_1,x_2,y_2}{} \ea
$$
is used to express that, at any step $t$ of the plan, a robot $r$
cannot move from $(x_1,y_1)$ to $(x_2,y_2)$ if there is no
collision-free trajectory between them. Here collision check is done
by the external predicate $\&\ii{path\_exists}$ implemented in C++,
utilizing the bidirectional RRT (Rapidly Exploring Random
Trees)~\cite{kuffnerrrt} as in the OMPL~\cite{ompl} library.

In addition to the classical negation, ASP considers another sort of negation: ``negation as failure'' (denoted $\no$, and often called ``default negation''). Intuitively,  $\neg p$ expresses that we know that $p$ does not hold, whereas $\no\ p$ expresses that we do not know that $p$ holds. This second sort of negation empowers ASP to express our assumptions (called ``defaults'') when we do not have complete knowledge. For instance, we can express that ``every object $o$ in a kitchen is assumed to be on the counter unless they are known to be on the table''  by the following ASP rule
$$
\ba l
\ii{at}(o,\ii{Counter},t) \lar \no\ \ii{at}(o,\ii{Table},t) .
\ea
$$

In ASP, we use special constructs to express nondeterministic choices, cardinality constraints, and
optimization statements. For instance, the following ASP rule (called a `choice rule'')
$$
\{ \ii{sense}(\ii{at}(o),t) \}
$$
contains the choice expression $\ii{sense}(\ii{at}(o),t)$ in the head. For every object~$o$ and time step~$t$, this choice expression describes a subset of the set $\{ \ii{sense}(\ii{at}(o),t) \}$. Therefore, this rule expresses that, for every object~$o$ and time step~$t$, the action of sensing that the location of~$o$ may occur at~$t$.

The following ASP constraint (called a ``cardinality constraint'')
$$
\lar 2\{\ii{atRob}(l,t): \ii{robloc}(l)\}
$$
contains the cardinality expression $2\{\ii{atRob}(l,t): \ii{robloc}(l)\}$ in the body. For every time step $t$, this expression describes subsets of the set $\{\ii{atRob}(l,t): \ii{robloc}(l)\}$ whose cardinality is at least 2. Therefore, this constraint is used to ensure that, for every time step~$t$, the robot cannot be at two different locations at~$t$.

The following ASP expression (called an ``optimization statement'')
\beq \#\ii{minimize}\ [\ \ii{cost}(r,c,t) : \ii{robot}(r) = c\ ]
\eeq{ex:mincost}
is used to minimize the sum of all costs $c$ of robotic actions
performed in a plan, where costs of actions performed by robot $r$
at time step $t$ are defined by atoms of the form
$\ii{cost}(r,c,t)$.

A version of external atoms (where predicate names are not passed as input arguments), and all the constructs described above are supported by the ASP solver \clingo\ used as part of \hcpasp. For more information about the input language of \clingo, we refer the reader to \clingo's manual: {\small\url{https://sourceforge.net/projects/potassco/files/guide/2.0/guide-2.0.pdf}}  (June 18, 2017).

%%%%%%%%%%%%%%%%%%%%%%%%%%%%%%%%%%%%%%%%%%%%%%%%%%%%%%%%%%%%%%%%%%%%%%%%%
\section{Hybrid Classical Planning in ASP}

Classical planning considers complete knowledge (under full observability) over a dynamic domain. A dynamic domain is defined by means of fluent constants and (actuation) action constants: A world state can be characterized by an interpretation of fluent constants, whereas an action is characterized by an interpretation of action constants. Then, dynamic domains under full observability can be modeled as transition systems --- directed graphs where nodes denote the world states of the domain, and edges denote the transitions between these states caused by occurrences or nonoccurrences of actions in that domain. Note that such transition systems respect the Markov principle (i.e., actions do not have delayed effects).

Given an initial state $s_0$, goal conditions $G$, and a nonnegative integer $k$, classical planning asks for a sequence $P=\langle a_0,a_1,...,a_{k-1}\rangle$ of actions, which characterizes a path $X=\langle s_0,s_1,...,s_{k-1},s_k\rangle$ from $s_0$ to a goal state $s_k$ in this transition system such that every edge $(s_i,s_{i+1})$ in the path characterizes an occurrence of action $a_i$. This sequence $P$ of actions is called a plan, with makespan $k$. The history~$H=\langle s_0,a_0,s_1,a_1,...,s_{k-1},a_{k-1},s_k\rangle$ of a plan describes the path $X$ by depicting also the relevant actions. Classical planning is NP-complete for polynomially bounded plans~\cite{ErolNS95}.

For robotic domains, to ensure executability of classical plans, e.g., to ensure continuous movements along collision-free trajectories, further checks need to be performed. This requires combining classical planning with feasibility checks. We call this problem hybrid classical planning. As discussed in Section~2 of the main paper, there are different methods of integrating planning with feasibility checks. We consider solving hybrid classical planning problems in ASP, using HEX programs, with respect to appropriate methods of integration~\cite{ErdemPS16}.

\paragraph{Representing hybrid action domains in ASP}

We formalize hybrid dynamic domains in ASP, under full observability and as a transition system, in the spirit of~\cite{ErdemGL16}. Such a description of a hybrid dynamic domain in ASP relies on three important forms of rules.

For a formula $H$ and an index $i$ (for time step), let us denote by
$H(i)$ the formula obtained from $H$ by replacing every atom $q$ by
$q(i)$. Intuitively, $H(i)$ expresses that the formula $H$ holds at
time step $i$.

\noindent \underline{Effect rules:}
Direct outcomes of actions are expressed with {\em effect rules} of the form
\beq E(i+1) \lar A(i), F(i) \eeq{eq:effect}
where $A$ is a conjunction of action atoms, $E$ is a fluent literal, and $F$ is conjunction of fluent literals. This rule indicates that if the actions in $A$ are executed at time step $i$ where $F$ holds then at the next state $E$ holds. For instance, the following effect rule describes an effect of a ``move'' action of a mobile robot $r$ navigating to a location $l$ at time step $i$:
$$% \beq
at(r,l,i+1) \lar move(r,l,i) .
$$% \eeq{eq:effect-exp}
It expresses that, as a direct effect of this action, the location of robot $r$ changes to $l$ at the next time step $i+1$.

\noindent \underline{Precondition rules:} Preconditions of actions are expressed with {\em precondition rules} of the form
\beq \lar A(i), F(i), \no\ G(i). \eeq{eq:precondition}
where $A$ is a conjunction of action atoms, and $F$ and $G$ are conjunctions of fluent literals.
The precondition rule above expresses that, to execute an action $A$ at time step $i$ at a state where $F$ holds, the action's preconditions $G$ must hold. For instance, according to the following precondition rule
$$% \beq
\lar move(r,l,i), at(r,l,i)
$$%\eeq{eq:precondition-exp}
action $move(r,l)$ is possible if the robot is not already at the destination location $l$.

\noindent \underline{Hybrid rules:}
A {\em hybrid rule} is a rule where the right hand side of $\lar$
includes external atoms. External atoms~\cite{hex2005} are not
fluent or action constants; their truth values are computed
externally (out of ASP).

These rules are important for robotics applications since low-level feasibility checks for each action can be computed externally and then integrated into transition system description by means of external atoms.
For instance, the following hybrid precondition rule ensures that, at time step $i$, a robot $r$ can move from its current location $x$ to its destination location $l$ if there is a collision-free trajectory between them:
$$%\beq
\lar at(r,x,i), move(r,l,i), \no\ \ext{move\_is\_feasible}{r,l,x}{} .
$$%\eeq{eq:hybrid-exp}
The external atom $\ext{move\_is\_feasible}{r,l,x}{}$ passes $r$, $l$, $x$ as inputs to the external computation (e.g., a Python program) that calls a motion planner to check the existence of a collision free trajectory for $r$ from $x$ to $l$, and then returns the result of the computation to the precondition rule.

\paragraph{Defining planning problems in ASP}

Once the domain is described in ASP, we can specify the initial state of the world in different ways, e.g., by a set of facts, like
$$
\ii{atRob}(\ii{Table},0) ,
$$
by choice rules accompanied with constraints, like
$$
\ba l
\{ \ii{atRob}(l,0) \}\\
\lar \no\ \ii{atRob}(\ii{Table},0) ,
\ea
$$
or by assumptions, like
$$
\ii{atRob}(\ii{Table},0) \lar \neg\ \ii{atRob}(\ii{Table},0) .
$$

We can describe the goal conditions by a set of rules, and add constraints to ensure their reachability, as in the examples below:
$$
\ba l
\ii{goal} \lar \ii{atRob}(\ii{Table},t) \\
\lar \no\ \ii{goal} .
\ea
$$

Then, with a domain description and planning problem description, plans of actuation actions can be computed using the ASP solver \clingo.

%%%%%%%%%%%%%%%%%%%%%%%%%%%%%%%%%%%%%%%%%%%%%%%%%%%%%%%%%%%%%%%%%%%%%
\section{An Example for Hybrid Conditional Planning: Kitchen Domain}

\begin{figure}[htb]
	\vspace*{-1\baselineskip}
\centering
\resizebox{0.8\columnwidth}{!}{\rotatebox{0}{\includegraphics{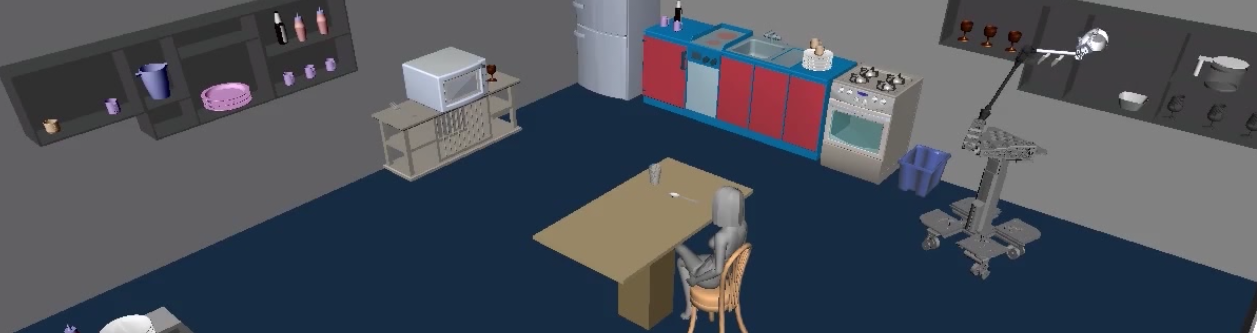}}}\vspace{0.125\baselineskip}
\resizebox{0.8\columnwidth}{!}{\rotatebox{0}{\includegraphics{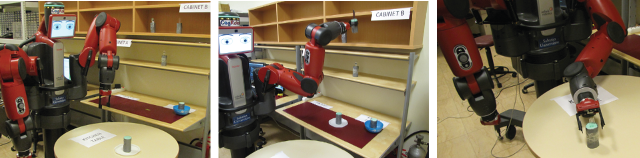}}}     \vspace{-1\baselineskip}
\caption{The robot is manipulating a bowl from Cabinet B in dynamic simulation (top) and a fork during physical implementation (bottom).}
\vspace{-1\baselineskip}
\label{fig:bowl-pick}
\end{figure}

As a benchmark for hybrid conditional planning, we consider a dynamic service robotics scenario, where a bimanual mobile manipulator is responsible for setting up a kitchen table, as depicted in Figure~\ref{fig:bowl-pick}. This domain is introduced in \cite{NoumanYEP16}: ``The mobile manipulator can navigate around the kitchen to  pick up and place objects as long as collision free trajectories exist. Kitchenware, such as mugs, spoons, knives, plates may be found in
cabinets or may be left on other flat surfaces, such as counter tops or shelves.  In the kitchen, there also exists a faucet to clean kitchenware as required. Finally, there is a kitchen table, where the proper kitchenware must be placed on to comply with table setting etiquette.''

In this domain, there are four actuation actions: $goto$, $pickup$, $placeon$ and $clean$.
For the feasibility of these actions, existence of a collision-free trajectory is implemented based on \ompl~\cite{ompl} (to be used as a precondition of $goto$ action), while reachability, graspability and inverse kinematics checks are implemented based on \openrave~\cite{diankov_thesis} (to be used as preconditions of $pickup$ and $placeon$ actions).

This domain contains three types of uncertainties. First, the person might have different
food preferences (e.g., soup, pizza, salad), which can only be revealed
when directly communicated with the user during plan execution.
Second, the locations of some kitchenware may not be known by the robot during the planning phase. These
locations can  be reliably gathered only if the robot actively
searches for these objects when it needs to use them. Third,
the cleanliness/dirtiness of the objects may not be known in advance for sure.
Along these lines, three sensing actions are considered: $checkFoodType$,
$checkLoc$ and $checkisClean$.

An ASP description of this domain presented to \hcpasp, in the input language of \clingo, is provided in Figures~\ref{fig:kitchen1}--\ref{fig:kitchen7}. The description consists of three parts, appropriate for incremental grounding (and thus incremental computation of plans), and preceded by the expressions {\small\tt \#program base}, {\small\tt \#program step(t)}, and {\small\tt \#program check(t)}. Intuitively, the first part describes the domain predicates and the general knowledge about the world at time step 0; so it is instantiated once. The second part describes the states at time step {\small\tt t} and transitions of the world for time steps {\small\tt t-1} and {\small\tt t}. Here, the value of {\small\tt t} increases one by one starting from {\small\tt 1} until a plan is found; so this part is instantiated incrementally. This incremental grounding guarantees finding a plan with a minimum length.  The direct/indirect effects of actions, inertia, and action occurrences are defined in the second part.  The third part describes all the constraints to be checked at every time step {\small \tt t}. Here, the value of {\small\tt t} also increases one by one starting from {\small\tt 0} until a plan is found; so this part is instantiated incrementally as well. The preconditions of actions, state constraints, transition constraints, concurrency constraints, and the goal are defined in the third part.

The actuation actions are defined as described in Appendix B, whereas the sensing actions are defined as described in Section~5 of the main paper.

\begin{figure}[htb]
{\small \begin{verbatim}
#include <incmode>.

% maximum plan length
#const step_limit=40.

#program base.

% robotic manipulators
manip(manip_R;manip_L).

% objects and their types
object(bowl_0). object(bowl_1). object(spoon_0).  object(spoon_1).
object(fork_0). object(fork_1).  object(knife_0). object(knife_1).
object(plate_0).  object(plate_1).  object(wineGlass_0).
object(wineGlass_1). object(waterGlass_0). object(waterGlass_1).

type(bowl,bowl_0). type(bowl,bowl_1). type(spoon,spoon_0).
type(spoon,spoon_1). type(fork,fork_0). type(fork,fork_1).
type(wineGlass,wineGlass_0). type(wineGlass,wineGlass_1).
type(knife,knife_0). type(knife,knife_1).
type(plate,plate_0). type(plate,plate_1).
type(waterGlass,waterGlass_0). type(waterGlass,waterGlass_1).

% number of objects in each type
type_T(T) :- type(T,O).
type_no(T,N) :- #count{O: type(T,O)}=N, type_T(T).

% possible locations of objects
objlocnotonhold(extratable; cabinetA; cabinetB; faucet; table).
objloc(L) :- objlocnotonhold(L).
objloc(M) :- manip(M).
#const objloc_size= 7.

% possible locations of the robot
robloc(L):- objlocnotonhold(L).

% food types
food(soup; pizza; chicken).

% which utensils are expected to be on the table for which food type
expected_T(soup,bowl). expected_T(soup,spoon).
expected_T(soup,waterGlass).
expected_T(pizza,fork). expected_T(pizza,knife).
expected_T(pizza,plate). expected_T(pizza,wineGlass).
expected_T(chicken,fork). expected_T(chicken,knife).
expected_T(chicken,plate). expected_T(chicken,bowl).
expected_T(chicken,wineGlass).
unexpected(F,O) :- type(T,O), not expected_T(F,T), food(F).
\end{verbatim} }
\caption{Kitchen table setting domain presented to \hcpasp, in the input language of \clingo: Part~1 -- Domain predicates.}
    \label{fig:kitchen1}
      \vspace{-.5\baselineskip}
\end{figure}

\begin{figure}[htb]
{\small \begin{verbatim}
% time_min = 0

% ramifications for t=0

% if an object is located at L1 then it is not anywhere else
-atObj(O,L,time_min) :- atObj(O,L1,time_min), object(O), objloc(L),
  objloc(L1), L!=L1.
% if a food type is requested, no other food type can be requested
-requested(F,time_min) :- requested(F1,time_min), food(F),
  food(F1), F!=F1.

% if an object's location is unknown then it cannot be on
%   the robot's hand either.
-atObj(O,M,time_min) :- {atObj(O,L,time_min):objloc(L)}0,
  manip(M), object(O).

% if an object O is not at any of the object locations (except L),
%   then it should be at L
atObj(O,L,time_min) :-
  objloc_size-1{-atObj(O,L1,time_min):objloc(L1),L1!=L}objloc_size-1,
  objloc(L), object(O).

% by default the objects are not on the table
-atObj(O,table,time_min) :- not atObj(O,table,time_min), object(O).

% actions are initially exogenous
{move(L,time_min)} :- robloc(L).
{pickUp(M,O,time_min)} :- manip(M), object(O).
{place(M,time_min)} :- manip(M).
{clean(M,time_min)} :- manip(M).
{sense(cleanObj(O),time_min)} :- object(O).
{sense(locObj(O),time_min)} :- object(O).
{sense(food_request,time_min)}.
\end{verbatim} }
\caption{Kitchen table setting domain presented to \hcpasp, in the input language of \clingo: Part~2 -- State constraints and possible action occurrences initially.}
    \label{fig:kitchen2}
      \vspace{-.5\baselineskip}
\end{figure}

\begin{figure}[htb]
{\small \begin{verbatim}
#program step(t).

% inertia for fully observed fluents
%   (with uniqueness and existence constraints)
{atRob(L,t+time_min)} :- atRob(L,t+time_min-1), robloc(L).

% inertia for partially observed fluents
atObj(O,L,t+time_min) :- not -atObj(O,L,t+time_min),
  atObj(O,L,t+time_min-1), object(O), objloc(L).
-atObj(O,L,t+time_min) :- not atObj(O,L,t+time_min),
  -atObj(O,L,t+time_min-1), object(O), objloc(L).

isclean(O,t+time_min) :- not -isclean(O,t+time_min),
  isclean(O,t+time_min-1), object(O).
-isclean(O,t+time_min) :- not isclean(O,t+time_min),
  -isclean(O,t+time_min-1), object(O).

requested(F,t+time_min) :- not -requested(F,t+time_min),
  requested(F,t+time_min-1), food(F).

% ramifications for t>0

% if an object is located at L1 then it is not anywhere else
-atObj(O,L,t+time_min) :- atObj(O,L1,t+time_min), object(O),
  objloc(L), objloc(L1), L!=L1.

% if a food type is requested, no other food type can be requested
-requested(F,t+time_min) :- requested(F1,t+time_min), food(F),
  food(F1), F!=F1.

% if an object's location is unknown then it cannot be on
%   the robot's hand either.
-atObj(O,M,t+time_min) :- {atObj(O,L,t+time_min):objloc(L)}0,
  manip(M), object(O).

% if an object O is not at any of the object locations (except L),
%   then it should be at L
atObj(O,L,t+time_min) :-
  objloc_size-1{-atObj(O,L1,t+time_min):objloc(L1),L1!=L}objloc_size-1,
  objloc(L), object(O).
\end{verbatim} }
\caption{Kitchen table setting domain presented to \hcpasp, in the input language of \clingo: Part~3 -- Inertia and ramifications.}
    \label{fig:kitchen3}
      \vspace{-.5\baselineskip}
\end{figure}

\begin{figure}[htb]
{\small \begin{verbatim}
% action occurrences
{move(L,t+time_min)} :- robloc(L).
{pickUp(M,O,t+time_min)} :- manip(M), object(O).
{place(M,t+time_min)} :- manip(M).
{clean(M,t+time_min)} :- manip(M).
{sense(cleanObj(O),t+time_min)} :- object(O).
{sense(locObj(O),t+time_min)} :- object(O).
{sense(food_request,t+time_min)}.

% direct effects of actions
% move(R,L)
atRob(L,t+time_min) :- move(L,t+time_min-1), robloc(L).

% pickUp(R,M,O)
atObj(O,M,t+time_min) :- pickUp(M,O,t+time_min-1),
  object(O), manip(M).

% place(robot,man,loc)
atObj(O,L,t+time_min) :- place(M,t+time_min-1),
  atObj(O,M,t+time_min-1), atRob(L,t+time_min-1),
  manip(M), object(O), objlocnotonhold(L).

% clean(R,M)
isclean(O,t+time_min) :- clean(M,t+time_min-1),
  atObj(O,M,t+time_min-1), manip(M), object(O).

% sensing actions
1{isclean(O,t+time_min);-isclean(O,t+time_min)}1 :-
  sense(cleanObj(O),t+time_min-1), object(O).
1{atObj(O,L,t+time_min):objlocnotonhold(L)}1 :-
  sense(locObj(O),t+time_min-1), object(O).
1{requested(F,t+time_min):food(F)}1 :-
  sense(food_request,t+time_min-1).

\end{verbatim} }
\caption{Kitchen table setting domain presented to \hcpasp, in the input language of \clingo: Part~4 -- Action occurrences and their direct effects.}
    \label{fig:kitchen4}
      \vspace{-.5\baselineskip}
\end{figure}

\begin{figure}[htb]
{\small \begin{verbatim}
#program check(t).

% uniqueness and existence constraints
:- 2{atRob(L,t+time_min): robloc(L)}.
:- {atRob(L,t+time_min): robloc(L)}0.

:- 2{atObj(O,M,t+time_min):object(O)}, manip(M).

% preconditions of actions
% the robot cannot move to L if it is already there
:- move(L,t+time_min), atRob(L,t+time_min), robloc(L).

% the robot cannot pick up an object
%  if it is already holding one
pickUpRM(M,t+time_min) :-  pickUp(M,O,t+time_min),
  object(O), manip(M).
:- pickUpRM(M,t+time_min), 1{atObj(O,M,t+time_min):object(O)},
  manip(M).

% the robot cannot pick up an object
%  if the object is not at the same place as robot.
pickUpRO(O,t+time_min) :-  pickUp(M,O,t+time_min),
  manip(M), object(O).
:- pickUpRO(O,t+time_min), not atRob(L,t+time_min),
  atObj(O,L,t+time_min), object(O), robloc(L).
:- pickUpRO(O,t+time_min), atRob(L,t+time_min),
  not atObj(O,L,t+time_min), object(O), robloc(L).

% the robot cannot place an object if it is not holding any
:- place(M,t+time_min), {atObj(O,M,t+time_min):object(O)}0,
  manip(M).

% the robot cannot clean if it is not at the faucet
cleanR(t+time_min) :- clean(M,t+time_min), manip(M).
:- cleanR(t+time_min), not atRob(faucet,t+time_min).
:- clean(M,t+time_min), {atObj(O,M,t+time_min):object(O)}0,
  manip(M).
:- clean(M,t+time_min), atObj(O,M,t+time_min),
  object(O), manip(M),
{isclean(O,t+time_min); -isclean(O,t+time_min)}0.

% sensing is not possible if the values of relevant
%  fluents are known
:- sense(cleanObj(O),t+time_min),
  1{isclean(O,t+time_min); -isclean(O,t+time_min)},
  object(O).
:- sense(cleanObj(O),t+time_min),
  {atObj(O,M,t+time_min): manip(M)}0, object(O).
:- sense(locObj(O),t+time_min),
  1{atObj(O,L,t+time_min): objloc(L)}.
:- sense(food_request,t+time_min),
  1{requested(F,t+time_min): food(F)}.
  \end{verbatim} }
\caption{Kitchen table setting domain presented to \hcpasp, in the input language of \clingo: Part~5 -- State constraints and preconditions of actions.}
    \label{fig:kitchen5}
      \vspace{-.5\baselineskip}
\end{figure}

\begin{figure}[htb]
{\small \begin{verbatim}
% feasibility checks

:- move(L1,t+time_min), atRob(L,t+time_min),
  robloc(L), robloc(L1), @move_feasible(L,L1)!=1.

:- pickUp(M,O,t+time_min), manip(M), object(O),
  atRob(L,t+time_min), atObj(O,L,t+time_min),
  robloc(L), @pickUp_feasible(M,O,L)!=1.

:- place(M,t+time_min), atObj(O,M,t+time_min),
  manip(M), object(O), atRob(L,t+time_min),
  robloc(L), @place_feasible(M,O,L)!=1.

% concurrency constraints

moveR(t+time_min) :- move(L,t+time_min), robloc(L).
pickUpR(t+time_min) :- pickUp(M,O,t+time_min), manip(M), object(O).
placeR(t+time_min) :- place(M,t+time_min), manip(M).

% the robot cannot pick/place/clean an object while moving
:- moveR(t+time_min), pickUpR(t+time_min).
:- moveR(t+time_min), placeR(t+time_min).
:- moveR(t+time_min), cleanR(t+time_min).

% the robot cannot pick/place an object while cleaning it
:- pickUpRM(M,t+time_min), clean(M,t+time_min), manip(M).
:- place(M,t+time_min), clean(M,t+time_min), manip(M).
:- place(M,t+time_min), pickUpRM(M,t+time_min), manip(M).

actAction(t+time_min):- moveR(t+time_min).
actAction(t+time_min):- pickUpR(t+time_min).
actAction(t+time_min):- placeR(t+time_min).
actAction(t+time_min):- cleanR(t+time_min).

sensAction(t+time_min):- sense(cleanObj(O), t+time_min), object(O).
sensAction(t+time_min):- sense(locObj(O), t+time_min), object(O).
sensAction(t+time_min):- sense(food_request, t+time_min).

% no sensing action and actuation action can occur
%  at the same time
:- actAction(t+time_min), sensAction(t+time_min).

% no two sensing actions are allowed at the same time
:- 2{sense(cleanObj(O),t+time_min):object(O);
     sense(locObj(O1),t+time_min):object(O1);
     sense(food_request,t+time_min)}.
\end{verbatim}}
\caption{Kitchen table setting domain presented to \hcpasp, in the input language of \clingo: Part~6 -- Feasibility checks and concurrency constraints.}
    \label{fig:kitchen6}
      \vspace{-.5\baselineskip}
\end{figure}

\begin{figure}[htb]
{\small \begin{verbatim}
% goal conditions

% what is expected on the table wrt food type
expected(F,T,t+time_min) :-
  1{atObj(O,table,t+time_min):type(T,O),isclean(O,t+time_min)}1,
  N-1{-atObj(O,table,t+time_min):type(T,O)}N-1,
  food(F), expected_T(F,T), type_no(T,N).

% the goal is not satisfied
%  if the expected objects are not on the table or
%  if there is an unexpected object on the table
notgoal(t+time_min) :- not expected(F,T,t+time_min),
  expected_T(F,T), requested(F,t+time_min), food(F).
notgoal(t+time_min) :- not -atObj(O,table,t+time_min),
  unexpected(F,O), object(O), requested(F,t+time_min), food(F).

% the goal is satisfied otherwise
goal(t+time_min) :- not notgoal(t+time_min),
  expected(F,_,t+time_min), requested(F,t+time_min), food(F).

% ensure that goal is reached some time before step_time
:- query(t), not goal(t+time_min).
\end{verbatim} }
\caption{Kitchen table setting domain presented to \hcpasp, in the input language of \clingo: Part~7 -- Goal conditions.}
    \label{fig:kitchen7}
      \vspace{-.5\baselineskip}
\end{figure}

%%%%%%%%%%%%%%%%%%%%%%%%%%%%%%%%%%%%%%%%%%%%%%%%%%%%%%%%%%%%%%%%%%%%%%%%%
\section{Experimental comparison of \hcpasp\ with \ascp} \label{sec:ascp}

As discussed in Section~2 of the main paper, although not compilation-based, the offline non-hybrid conditional planner \ascp\ also uses ASP to compute conditional plans. Therefore, we have compared these two ASP-based conditional planners, over one of the benchmarks of \ascp: Bomb in the Toilet with Sensing Actions (BTS)~\cite{WeldAS98}. In this domain, it has been alarmed that there is a bomb in the toilet. There are $m$ suspicious packages, and one of them contains the bomb.  The bomb can be defused by dunking the package with bomb into the toilet; dunking a package clogs the toilet and flushing the toilet unclogs it.  The existence of a bomb in a package can be sensed by a metal detector, by a dog to sniff the bomb, or by an x-ray machine. Initially, the bomb is armed and the toilet is not clogged; the goal is that the bomb is disarmed and the toilet is not clogged.

We have experimented with \ascp\ using the ASP encoding of BTS ({\tt bt\_3sa.smo})\footnote{ \url{https://www.cs.nmsu.edu/~tson/ASPlan/Sensing/test/bt_3sa.smo}.} transformed from ${\cal A}^c_K$ with the ASP solver \clingo. The results of experiments for $m=10,11, .., 17$ are shown in Table~\ref{tab:ascp}; the computation time for \ascp\ does not contain the time for transformation.

According to these results, finding a tree with one call of \clingo\ (using \ascp) takes more time, compared to computing and combining the branches of the tree in parallel (using \hcpasp).  For instance, for $m=17$, it takes more than an hour to compute a tree with \ascp\, whereas it takes about a second for \hcpasp.

\begin{table}[htb]
\centering \small
\vspace{-1.25\baselineskip}
\setlength{\tabcolsep}{0.5pt}
\caption{Comparison of \ascp\ with \hcpasp.}
\vspace{-.2\baselineskip}
\label{tab:ascp}
\begin{tabular}{|c|c|c|r|r|}
  \cline{1-5}
  No of Package  & Max Depth & Tree Size & \ascp\ Time & \hcpasp\ Time \\
                 &  &  & [sec] & (parallel with 20 threads) [sec]\\
  \cline{1-5}
  10 & 10 & 19 & 21   & 0.4 \\
  11 & 11 & 21 & 21   & 0.5 \\
  12 & 12 & 23 & 46   & 0.6 \\
  13 & 13 & 25 & 433  & 0.5 \\
  14 & 14 & 27 & 406  & 0.7 \\
  15 & 15 & 29 & 1953 & 0.9 \\
  16 & 16 & 31 & 2896 & 0.8 \\
  17 & 17 & 33 & 5807 & 1.0 \\
  \cline{1-5}
\end{tabular} \normalsize
\vspace{-1\baselineskip}
\end{table}

\end{appendix}

\vspace{-2ex}
\bibliographystyle{acmtrans}
%\bibliography{references}

%%%%%%%%%%%%%%%%%%%%%%%%%%%%%%%%%%%%%%%%%%%%%%%%%%%%%%%%%%%%

\label{lastpage}
\end{document}